\documentclass[twoside,11pt]{article}

\usepackage{jmlr2e}
\usepackage{natbib}

\usepackage[utf8]{inputenc} 
\usepackage[T1]{fontenc}    
\usepackage{hyperref}       
\usepackage{url}            
\usepackage{booktabs}       
\usepackage{amsfonts}       
\usepackage{amsmath}
\usepackage{multirow}
\usepackage{algorithm}
\usepackage{xcolor}
\newcommand{\bfs}[1]{{\boldsymbol #1}}

\newcommand{\R}{\mathcal{R}}
\newcommand{\E}{\mathbb{E}}
\newcommand{\F}{\mathcal{F}}
\newcommand{\I}{\mathcal{I}}

\newcommand{\D}{\mathcal{D}}
\newcommand{\Loss}{\mathcal{L}}

\newcommand{\rev}[1]{{\color{black} #1}}

\usepackage{algpseudocode}
\usepackage{nicefrac}       
\usepackage{microtype}      

\usepackage{graphicx}
\usepackage{logicproof}
\usepackage{subcaption}
\graphicspath{ {Figures/} }

\begin{document}

\title{Practical Attribution Guidance for Rashomon Sets}

\author{\name Sichao Li \email sichao.li@anu.edu.au \\
       \addr School of Computing\\
       Australian National University\\
       Canberra, ACT 2601, AU
       \AND
       \name Amanda S. Barnard \email amanda.s.barnard@anu.edu.au \\
       \addr School of Computing\\
       Australian National University\\
       Canberra, ACT 2601, AU
       \AND
       \name Quanling Deng \email quanling.deng@anu.edu.au \\
       \addr School of Computing\\
       Australian National University\\
       Canberra, ACT 2601, AU
}
       

\maketitle

\begin{abstract}
Different prediction models might perform equally well (Rashomon set) in the same task, but offer conflicting interpretations and conclusions about the data. The Rashomon effect in the context of Explainable AI (XAI) has been recognized as a critical factor. Although the Rashomon set has been introduced and studied in various contexts, its practical application is at its infancy stage and lacks adequate guidance and evaluation.
We study the problem of the Rashomon set sampling from a practical viewpoint and identify two fundamental axioms - generalizability and implementation sparsity that exploring methods ought to satisfy in practical usage.
These two axioms are not satisfied by most known attribution methods, which we consider to be a fundamental weakness. We use the norms to guide the design of an $\epsilon$-subgradient-based sampling method. We apply this method to a fundamental mathematical problem as a proof of concept and to a set of practical datasets to demonstrate its ability compared with existing sampling methods.
\end{abstract}


\section{Introduction}\label{sec:introduction}
Understanding the behavior of machine learning models is receiving considerable attention. 
Many researchers seek to identify important features and describe their effects depending on a specific model. 
Recently, researchers have argued that explaining a single model is insufficient and we should explain a set of similar-performing models. This set of models is called a Rashomon set \citep{fisher2019all}. A commonly agreed definition of the Rashomon set in the machine learning community is when the benchmark model $f^* \in \F$ minimizes the loss function, i.e.,
$\Loss(f^*) = \inf_{f\in \F} \Loss(f)$, the set defined by hypothesis space $\mathcal{H}$ on the basis of $\epsilon > 0$ is called the Rashomon set; see for example in \citep{dong2020exploring, NEURIPS2022_5afaa8b4, li2022variance}.
\begin{equation} \label{eq:grs}
\R(\epsilon, f^{*}, \F) = \{ f \in \F: \Loss(f(\bfs{X}), \bfs{y}) \leq (1 + \epsilon) \Loss(f^*(\bfs{X}), \bfs{y}) \}.
\end{equation}   

A significant challenge in exploring the whole Rashomon set in practice is that it is infeasible to enumerate all possible models \citep{semenova2022existence}. For example, the possible number of trees of depth at low as 4 with 10 binary features is more than $\text{9.338} \times \text{10}^{20}$ and the number of neural networks (NNs) in hypothesis space grows exponentially with the number of parameters \citep{hu2019optimal}. 
The recent work of \cite{hsu2022rashomon} studied the Rashomon ratio and the pattern Rashomon ratio 
to estimate the volume of the Rashomon set. This is a level set estimation problem and represents the fraction of models in the hypothesis space that fit the data about equally well. These theoretical results, such as the Rashomon ratio, can be helpful in explaining some phenomena observed in practice, while it is still a measure of a learning problem's complexity. For instance, Rashomon ratios tend to be
large for lower complexity function classes but are not necessarily intuitive \citep{semenova2024path}.
We aim to analyze the Rashomon set obtained in practice. 

Retraining models is impractical to explore the model space due to limited memory and computational resources. The mainstream of exploring the Rashomon set is sampling or approximating, acquiring a subset of the Rashomon set \citep{hsu2022rashomon}. 
\citep{dong2020exploring} provided a simple example of enumerating similarly-performing decision trees under certain settings, but not general for all cases. 
Thus, researchers usually define a reference model and sample a class of reference models to represent the whole Rashomon set. 
\citet{NEURIPS2022_5afaa8b4} recently provided a practical tree-based solution for a nonlinear discrete model class by fitting an optimal tree and exploring the whole Rashomon set of sparse decision trees.
\citet{dong2020exploring} also trained a logistic regressor and provided an ellipsoid approximation for the logistic model class. For NN-based models, \citet{li2022variance} provided a variance tolerance factor (VTF) to interpret all NNs by greedy searching an extra layer on the top of the network.  

\begin{figure}
    \centering
    \includegraphics[width=\textwidth]{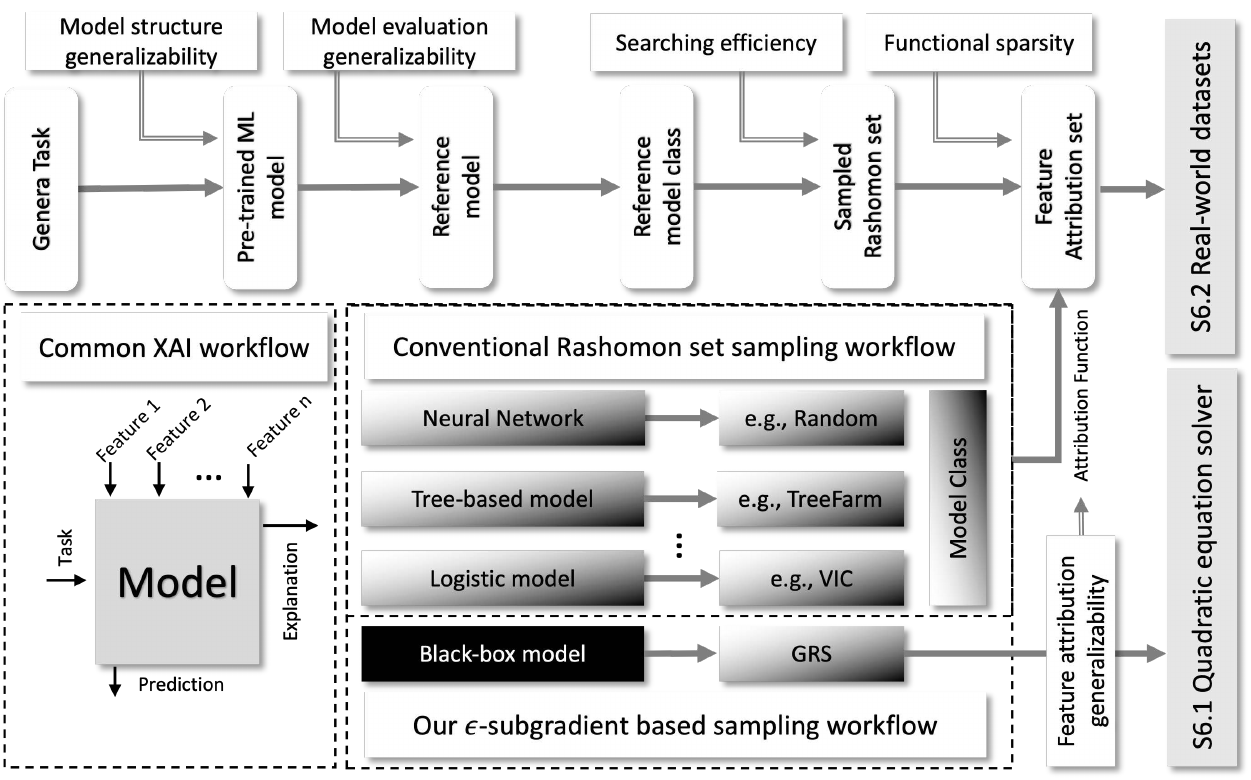}
    \caption{The pipeline at the top illustrates the overall Rashomon sampling process, highlighting the desired axioms at the corresponding positions. Detailed explanations are provided in the bottom dotted boxes. The first dotted box at the left shows the common XAI workflow, where researchers use a single ML model to derive explanations. The box at the top right outlines the conventional Rashomon set sampling workflow, where the structure of the reference model is fixed and assumed with prior knowledge, and then a model class is explored according to existing methods to approximate the Rashomon set. Here we use grayscale gradients to illustrate varying levels of interpretability among different ML structures. For example, a decision tree is generally interpretable, whereas an ensemble of trees tends to be less so. 
    In practice, a pre-trained model is often provided as a black-box model, extended in the bottom right box, which could include NNs, decision trees, and other ML models identified for the same task. Therefore, a general workflow is proposed and tested in various experimental settings in Sec. \ref{sec:experiments}.}
    \label{fig:work_flow}
\end{figure}

However, this pipeline assumes the prior knowledge/controllability of the reference model, e.g., linear model, tree-based model and generalized additive models (GAM) \citep{chen2023understanding}, a requirement which is not always satisfied in practice. Here we focus on a general context that the reference model is given as a black-box model that could be any pre-specified machine learning model under various learning settings. The conventional framework might suffer from the problem-setting and a more general workflow is needed, as shown in Fig. \ref{fig:work_flow}. Additionally, the lack of quantitative comparison of sampling methods hinders the potential development of the Rashomon set.
To compensate for these shortcomings, we take an axiomatic approach and identify two axioms from a practical perspective, namely generalizability and implementation sparsity, that Rashmon set sampling methods ought to consider when designing and employing in practice. Unfortunately, most previous methods fail to satisfy one of these two axioms.

The rest of the paper is organized as follows: 
In Sec. \ref{sec:notation}, we introduced common notations and terminologies used throughout this paper for readers to facilitate reader comprehension.
In Sec. \ref{sec:axioms}, we identified two main axioms and several sub-axioms for the Rashomon set sampling process in practice from existing literature. Specifically, we discussed generalizability in Sec. \ref{subsec:generalizability}, encompassing model structure generalizability, model evaluation generalizability, and feature attribution generalizability and explored implementation sparsity in Sec. \ref{subsec:implementation-sparsity}, including searching efficiency and functional sparsity.
Sec. \ref{sec:method} introduces an $\epsilon$-subgradient-based sampling framework, guided by the discovered axioms, where we introduced a generalized representation of the Rashomon set, and formalized a general feature attribution function. The relevant theories and proof are provided in both Sec. \ref{sec:method} and Sec. \ref{sec:practical-axioms}. 
In Sec. \ref{sec:experiments}, we present both synthetic and empirical studies to demonstrate our arguments and provide a visualized comparison of our method with other proposed approaches. 
We highlight two key findings here, namely: (1) Considering a set of models instead of a single model can be helpful in approaching the ``ground truth'' feature attributions, discussed in \ref{subsec:generalizability-in-syntheticdata} and (2) A potential trend of attributions exists across sampled Rashomon sets, shown in Sec. \ref{subsec:gene-stat}.
Additionally, in Sec. \ref{sec:conclusion}, we provide a general conclusion of our paper by discussing its advantages, limitations, and broader impacts on other fields.

\section{Notations and Terminologies}\label{sec:notation}

In this study, we use bold lowercase letters such as $\bfs{v}$ to represent a vector and ${v}_i$ denotes its $i$-th element.
Let the bold uppercase letters such as $\bfs{A}$ denote a matrix with ${a}_{[i, j]}$ being its $i$-th row and $j$-th column entry. 
The vectors $\bfs{a}_{[i, \cdot]}$ and $\bfs{a}_{[\cdot, j]}$ are its $i$-th row and $j$-th column, respectively.
\rev{Let $(\bfs{X}, \bfs{y})\in (\mathbb{R}^{n\times p}, \mathbb{R}^m)$ denote the dataset where $\bfs{X} = [\bfs{x}_{[\cdot, 1]},  \bfs{x}_{[\cdot, 2]}, \ldots, \bfs{x}_{[\cdot, p]}]$ is a $n \times p$ covariate input matrix and $ \bfs{y} $ is a $m$-length output vector.} We assume the observations are drawn i.i.d. from an unknown distribution $\D$.

Let $\I$ be a subset of feature indices: $\I \subset \{1, 2, \ldots, p \}$ and its cardinality is denoted by $|\I|$. 
All possible subsets are referred as $\mathbb{I} = \{\I \mid \I \subset \{1, 2, \ldots, p\} \}$. 
In the context of no ambiguity, we drop the square brackets on the vector and simply use $\bfs{x}_{s}$ to denote the feature. 
$\bfs{X}_{\setminus \bfs{s}}$ is the input matrix when the feature of interest (denoted as $\bfs{s}$ here) is replaced by an independent variable. 
Let $f :  \mathbb{R}^{n\times p} \to  \mathbb{R}^m$ be a predictive model and a model class is $\F \subset \{f \mid f\in \F\}$. $\Loss :  (f(\bfs{X}), \bfs{y}) \to \mathbb{R}$ be the loss function. 
The expected loss and empirical loss are defined as $L_{exp} = \E[L(f(\bfs{X}), \bfs{y})]$ and $L_{emp} = \rev{\frac{1}{n}} \sum_{i=1}^{n}L(f(\bfs{x}_{[i, \cdot]}), {y}_i)$, respectively. 

We denote $f_{ref}$ as the Rashomon set's reference or baseline model. The term ``feature attribution'' is used to denote the general contribution of the feature to the prediction.
Higher-order feature attribution refers to interactions among features. For clarity, we define first-order attribution as feature importance, second-order attribution as the interaction effect between two features, and so forth. Sampling and exploring are used for the same meaning in the context.

\section{Two Fundamental Practical Axioms}\label{sec:axioms}
\begin{figure}
    \begin{subfigure}[t]{0.25\textwidth}
        \centering
        \includegraphics[height=1.5in]{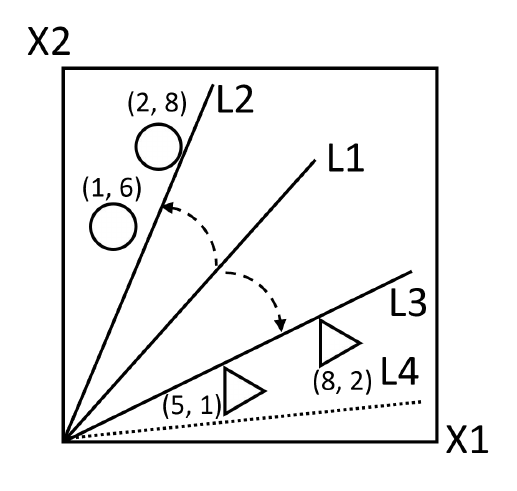}
        \caption{Linear model}
    \end{subfigure}%
        \begin{subfigure}[t]{0.25\textwidth}
        \centering
        \includegraphics[height=1.5in]{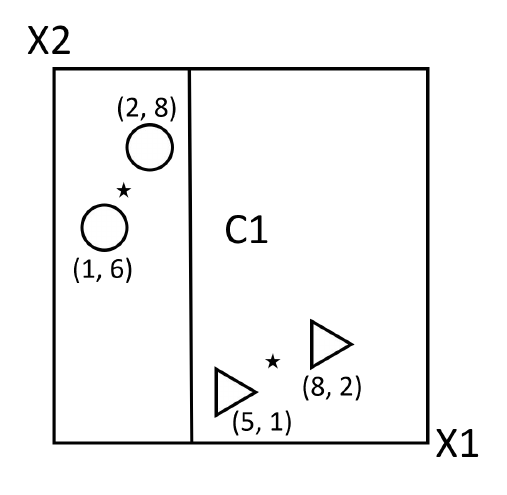}
        \caption{Clustering model}
    \end{subfigure}%
        \begin{subfigure}[t]{0.25\textwidth}
        \centering
        \includegraphics[height=1.5in]{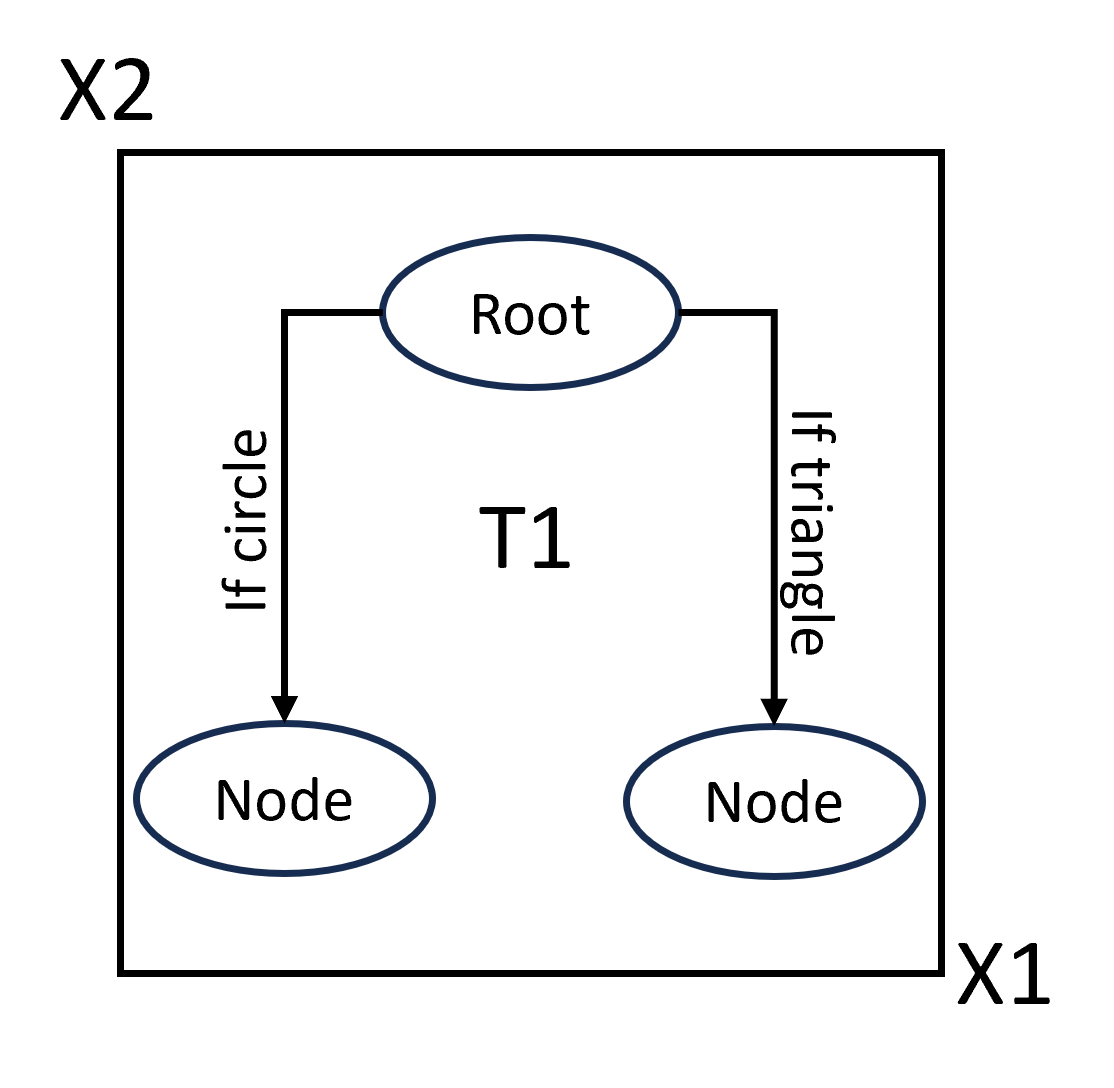}
        \caption{Decision tree}
    \end{subfigure}%
        \begin{subfigure}[t]{0.25\textwidth}
        \centering
        \includegraphics[height=1.5in]{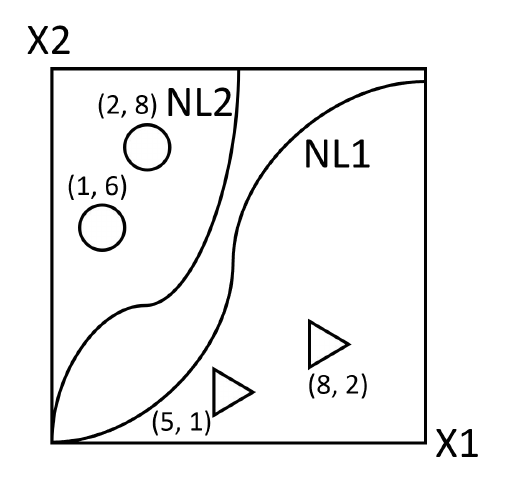}
        \caption{Non-linear model}
    \end{subfigure}%
    \caption{\rev{An example of binary classification. Each object is characterized by features $x$1, $x$2, and shape.} Possible models in the Rashomon set depend on different features for a simple circle and triangle classification. }
    \label{fig:toy-example}
\end{figure}

We now discuss two desirable characteristics when exploring the Rashomon set in practice. We find that other exploration methods in the literature break at least one of these axioms. As we will see in Sec. \ref{sec:method}, these axioms also guide the design of our method. To illustrate the idea behind axioms, \rev{we present an example of in-sample binary classification in Fig. \ref{fig:toy-example} and discuss the axioms below.}



\subsection{Axiom: Generalizability}\label{subsec:generalizability}
The core of the definition of Rashomon set is to find all near-optimal accurate models in the hypothesis space. Based on our observations in Sec. \ref{sec:introduction}, we first introduce \textit{model structure generalizability} and \textit{model evaluation generalizability}.

\subsubsection{Model Structure Generalizability}\label{subsubsec:ms-generalizability}
As a simple example \rev{shown in Fig. \ref{fig:toy-example}, the solution to the classification problem} can be discovered through multiple learning algorithms, e.g. supervised learning or unsupervised learning.  It can be accurately solved by linear classifiers (see Fig. \ref{fig:toy-example} (a)), and any hyperplane within the maximized margin can be considered acceptable in terms of accuracy. Non-linear classifiers can be defined in such a simple case, potentially requiring extra computational efforts (see Fig. \ref{fig:toy-example} (d)). Unsupervised clustering methods and decision trees can also achieve good accuracy (see Fig. \ref{fig:toy-example} (b) and (c)). The ideal sampling method is suitable for any type of baseline model, regardless of learning type and model structure, while most existing methods require training a specific model, e.g., tree-based structures in the work of \citep{NEURIPS2022_5afaa8b4}, or NNs in \citep{hsu2022rashomon}.

The selection of $f_{ref}$ is necessary to serve as a performance benchmark and it can be selected in different ways, such as minimizing the in-sample loss or sample splitting.  In some cases, it may be desirable to fix $f_{ref}$ equal to the best-in-class model $f^{*}$, but this is generally infeasible because $f^{*}$ is unknown \citep{fisher2019all}. For any $f_{ref} \in \F$, the Rashomon set $\R(f_{ref}, \epsilon, \F)$ will always be conservative in the sense that it contains the Rashomon set $\R(f^{*}, \epsilon, \F)$. In practice, we set $f_{ref}$ as a black-box model with $\Loss(f_{ref}) \geq \Loss(f^{*})$ and define the Rashomon set by $f_{ref}$. The sampled Rashomon set in practice is denoted by $\hat{\R}(f_{ref}, \epsilon, \F)$.
Here, we mainly focus on supervised, unsupervised, and semi-supervised learning approaches for a specific task. We illustrate the generalizability of the model structure through a synthetic problem with known ground truth in Sec. \ref{sec:experiments}.

\subsubsection{Model Evaluation Generalizability}\label{subsubsec:me-generalizability}

It is possible to train a model using different loss functions, such as mean squared error (MSE), Mean Absolute Error (MAE) for regression or log-loss, and ROC-AUC for classification. This is important because similar loss values can result in different decisions, e.g. [0.51, 0.49] versus [0.49, 0.51]. To ensure fairness in the Rashomon set, every sampled model will be evaluated using the same metric, regardless of its training type. For example, an ML model might be trained using log-loss in a classification task, but here we only consider the accuracy in the above case and thus all models are 100\% accurate. This property is satisfied by most sampling methods, as they mainly discuss the same model class.

\begin{remark}
    The evaluation matrices affect the number of possible models, as all the above models are performing well in terms of accuracy, while their training loss, e.g. Euclidean distance-based loss, may vary.  
\end{remark}

\subsubsection{Feature Attribution Generalizability}\label{subsubsec:fa-generalizability}

One of the main purposes of the Rashomon set exploration is interpretation/explanation \citep{rudin2019stop}. These two terms are used interchangeably in this paper. We mainly focus on feature interpretation, as discussed in previous works \citep{nauta2023anecdotal,zhong2022explainable, gola2018advanced, pankajakshan2017machine, 9007737}.
It is noted there are different attribution methods in the literature \citep{gilpin2018explaining, guidotti2018survey, adadi2018peeking, linardatos2020explainable, zhong2022explainable, roscher2020explainable, imrie2023multiple} and their attributions to the same feature might vary. For example, a task modeled by a linear neural network and a decision tree leads to two different explanations \citep{sundararajan2017axiomatic}, because the model architecture use the features differently. Considering the amount of interpretation methods in the current literature, and the lack of an interpretation evaluation benchmark, here we require the attribution method to be consistent in the sampling process so as to be comparable and the method should be generally applicable to offer higher-order attributions. Most methods only consider the first-order feature attribution and we formalize a loss-based generalized feature attribution in Sec. \ref{sec:method}.

\subsection{Axiom: Implementation Sparsity}\label{subsec:implementation-sparsity}
In this section, we focus on the sampling process and identify the axiom implementation sparsity, including searching efficiency and functional sparsity. 

\subsubsection{Searching Efficiency}\label{subsubsec:se-implementation-sparsity}
To ensure sampling efficiency in practice, a method should avoid training extra models that exceed the boundary to the maximum extent. As observed, there exists an infinite number of linear models in Fig. \ref{fig:toy-example} (a), some of which might misclassify objects, such as L4 in Fig. \ref{fig:toy-example} (a), and should be avoided during sampling. With more than $\text{9.338} \times \text{10}^{20}$ possible number of trees mentioned in Sec. \ref{sec:introduction}, it is impractical to enumerate all trees and subsequently filter valid ones due to the computational intensity. \citet{NEURIPS2022_5afaa8b4} proposed a top-down searching algorithm and significantly reduced the computational time. Similarly, the ellipsoid estimation saves computational time by searching from inner to outer \citep{dong2020exploring}. Random initialization of weights in networks is thus less computationally efficient. 


\subsubsection{Functional Sparsity}\label{subsubsec:fs-implementation-sparsity}
Two models are said to be \textit{functionally redundant} in the Rashomon set if their attribution outputs are the same for all features. It is desirable for sampling methods to satisfy the \textit{functional sparsity} axiom for explainability of the sampled model on any-order feature attributions, such as sparse attributions of features or interactions. This is because the purpose of exploring the Rashomon set is to provide more comprehensive attributions, from a single value to a range \citep{rudin2019stop}.
If two models are functionally redundant, indicating the explainability of the feature or feature set is constant, then one of these models should be avoided during sampling. 

This definition does not include implementation details. 
To show the sparsity of the sampled models, the sampling method should illustrate different model implementations with the same explanations from at least one attribution method.
However, two models that are implemented differently can provide the same explanations using gradients in NNs \citep{samek2016evaluating, pmlrv70shrikumar17a, tsang2017detecting}. Gradients are invariant to implementations based on the chain rule, e.g. $g$ and $f$ are the input and output of a function, respectively, and $h$ is the intermediate function, bu the gradient of input $g$ to output $f$ can be computed by 
$\frac{\partial f}{\partial g} = \frac{\partial f}{\partial h} \cdot \frac{\partial h}{\partial g}$. Either way, directly computing $\frac{\partial f}{\partial g}$ or invoking the chain rule via $h$, results in the same outcome.


\section{General Rashomon Subset Sampling Framework}\label{sec:method}
In this section, we outline an $\epsilon$-subgradient-based sampling framework guided by the proposed axioms, named the General Rashomon Subset Sampling (E-GRSS) framework, abbreviated as GRS.
We first introduce the generalized Rashomon set for practical use based on model structure and model evaluation generalizability due to the observations in practice that, for certain data sets and machine learning problems, finding this $f^*$ that minimizes the loss is extremely challenging or impossible, e.g. non-convex optimization and NP-hard problems \citep{lorenz1995essence, papadimitriou1998combinatorial}. 
\rev{Instead, we formalize} a generalized
feature attribution measurement with some useful statistics to quantify the Rashomon set.

\subsection{Generalized Representation for Models in the Rashomon Set}

\begin{definition}[generalized Rashomon set]
\label{def:grs}
Given any trained machine learning model $f_{ref}$, e.g. tree models, NNs, for any task, e.g. supervised learning, unsupervised learning\footnote{For unsupervised tasks, the training dataset contains $\{(\bfs{x}_i\}^{n}_{i=1}\}$ only. In that case, we set $f_{ref}(\bfs{x})$ as the ground truth output, and the baseline loss is 0. Other models are evaluated by $\Loss_{ref}(f(\bfs{x}), f_{ref}(\bfs{x}))$}, we assume this model is the ``optimal'' model for a certain task with a non-zero loss. Model optimization is not our focus in this study. The reference loss function remains the same for sampling models and this guarantees model evaluation generalizability.
Similar to previous work, the generalized Rashomon set is defined on the basis of a tolerance $\epsilon>0$ as below and the threshold for the Rashomon set is $\theta^{*}  = \epsilon \Loss_{ref}(f_{ref})$:
\begin{equation} \label{eq:grs}
\R(\epsilon, f_{ref}, \F) = \{ f \in \F: \Loss_{ref}(f(\bfs{X}), \bfs{y}) \leq (1 + \epsilon) \Loss_{ref}(f_{ref}(\bfs{X}), \bfs{y}) \}.
\end{equation}
\end{definition}

\rev{In practice, researchers sample from the generalized Rashomon set in different ways. Intuitively, we have the following. 
\begin{theorem}
    An empirical generalized Rashomon set is a subset of the generalized Rashomon set: $\hat{\R}(\epsilon, f_{ref}, \F) \subseteq \R(\epsilon, f_{ref}, \F)$. 
\end{theorem}
\noindent
\textbf{Proof. } The $\hat{\R}(\epsilon, f_{ref}, \F)$ at least contains the reference model
and the maximum set is when all possible models in $\R(\epsilon, f_{ref}, \F)$ are sampled. 
}

\begin{remark}
    Theoretically, there could exist a model that has a minimum loss of zero. However, the $\Loss_{ref}(f^*)=0$ for a trained model may not be of interest in terms of Eq. \eqref{eq:grs}.
    Therefore, we can define the generalized Rashomon set with a tolerance on the output labels as Eq. \eqref{eq:grs_zero} and the threshold $\theta^*$ becomes $\epsilon$.
    \begin{equation}\label{eq:grs_zero}
    \R(\epsilon, f_{ref}, \F) = \{ f \in \F: \Loss_{ref}(f) \leq \Loss_{ref}(f_{ref}) + \epsilon \}.
    \end{equation}
\end{remark}

\subsubsection{Input-activated characterization}
To characterize the generalized Rashomon set and find a generalized representation for models in the Rashomon set, we adopt the idea that every model in the Rashomon set can be replaced by the reference model.  This can be achieved by concatenating mask layers from \citep{li2023exploring}\footnote{We acknowledge the potential increasing complexity of the approximation, but this is not our focus.}, as all machine learning models can be accurately represented or approximated by NNs.  For example,  decision trees can be represented as NNs \citep{yang2018deep, Hinton2017Distilling, aytekin2022neural}; support vector machines as a shallow neural network; and K-means clustering with NNs \citep{sitompul2018optimization}. 

\begin{proposition}\label{prop:general_rs}
    The generalized Rashomon set can be represented as
    \begin{equation}\label{eq:grs_tau}
    \R(\epsilon, f_{ref}, \F) = \{ \bfs{Z} \in \mathbb{R}^{n \times p}: \Loss_{ref}(f_{ref}(\bfs{Z}, \bfs{y})) \leq \Loss_{ref}(f_{ref}(\bfs{X}, \bfs{y})) + \epsilon \},
    \end{equation}
\end{proposition}
\noindent where $\bfs{Z}$ can be seen as a transformation of $\bfs{X}$, formalized as $\bfs{Z} = m(\bfs{X})$, so that the model structure can remain the same in the sampling process. The property ensures model structure generalizability.

\subsubsection{Output-activated characterization}
Similarly, another idea is to reuse the reference model $f_{ref}$ and add additional layers to the output, since the output vector generally has components that have smaller point-wise errors than its other components. This affects the model's performance, but for simplicity, one may impose a zero bias for the layer. 
Alternatively, one may set the weights $\omega_{jk} = \delta_{jk}$ where $\delta_{jk}$ is the Kronecker delta ($\delta_{jk} = 1$ if $j=k$ and zero otherwise) and generalise the bias of the layer to a vector of biases for each component of the output vector. The idea is useful in instance-level Rashomon set and a similar work has been proved in \citep{hsu2022rashomon}, though our focus is feature-level explainability. 



\subsection{Generalized Feature Attribution With Useful Statistics}\label{subsec:gene-stat}

\subsubsection{generalized feature attribution formulation}
A general way of measuring feature importance, regardless of quantifying methods, can be defined as the conditional expected score of the feature to the specific feature value \citep{li2022variance, zien2009}. Here we adopt and extend the idea further for higher-order feature attributions in the Rashomon set. 

\begin{definition}[generalized feature attribution]
A pre-defined score function $s(\cdot)$ measures the importance of any order of features, e.g. $|\I|=1$ for feature importance and $|\I|>1$ for interaction importance, on the given data set. The conditional expected general feature attribution of $s_{\I}$ is the expected score $q_{\I}$ conditional to the feature set $\I$ of the model $f \in \F$, written as: 
\begin{equation}\label{eq:gfa_fn}
q_{\I}(f) = \mathbb{E} [ s_{\I}(\bfs{X}, \bfs{y}) \mid f \in \F ].
\end{equation}
\end{definition}

\begin{definition}[Feature attribution set and feature attribution space]\label{def:fat}
    A single model can attribute to all possible subset of features $\bfs{q}_{\mathbb{I}}(f) = \{\I \in \mathbb{I} : q_{\I}(f) \}$. Inversely, for a specific feature attribution, we can find a set of scores from all possible models $\bfs{q}_{f}(\I) = \{f \in \F : q_{\I}(f) \}$. Putting them together, we define a feature attribution space from the sampled Rashomon set as: $$\bfs{Q}_{\mathbb{I}}(\hat{\R}) = \{ q_{\I}(f) \mid f \in \hat{\R} ,  \I \in \mathbb{I} \}.$$
\end{definition}

\begin{theorem}
    A sampled Rashomon set always corresponds to a nonempty and finite feature attribution set.
\end{theorem}
\noindent
\textbf{Proof. } The minimum $\bfs{Q}_{\mathbb{I}}(\hat{\R})$ is when $\hat{\R} = \{f_{ref}\}$ and in practice, we only sample finite models from the hypothesis space. 


\subsubsection{Generalized feature attribution score function}
A general feature attribution score function can be used to calculate any order of feature importance, e.g. first-order feature importance, second-order feature interaction, and third-order feature impact. One common model-agnostic feature attribution measurement is a permutation-based method and it has been adopted as model reliance \citep{fisher2019all} and feature interaction score \citep{li2023exploring}. Here we apply the approach and adopt the score function for Eq. \eqref{eq:gfa_fn} as:
\begin{equation}
s_{\I}(\bfs{X}, \bfs{y}) = \left\{
\begin{matrix}
\varphi_{i}(\bfs{X}, \bfs{y}) \text{ if } |\I| = 1
\\ \nonumber
\varphi_{\I}(\bfs{X}, \bfs{y})-\sum_{i \in \I}\varphi_{i}(\bfs{X}, \bfs{y}) \text{ if } |\I| > 1
\end{matrix}\right.
\label{eq:fis_definition}
\end{equation}
where $\varphi_{\I}(\bfs{X}, \bfs{y}) =  \E[\Loss(f(\bfs{X}_{\setminus \I}), \bfs{y})] - \E[\Loss(f(\bfs{X}), \bfs{y})]$ is a measurement of feature attribution and $ \E[L(f(\bfs{X}), \bfs{y})]$ is the baseline effect that provides interpretability. We estimate the standard empirical loss by 
$\Loss_{emp}(f(\bfs{X}), \bfs{y}) = \frac{1}{n}\sum_{i=1}^{n}\Loss(f(\bfs{x}_{[i,\cdot]}), {y}_i)$ 
and permutating all possible combinations of the observed values to estimate: $$\Loss_{emp}(f(\bfs{X}_{\setminus i}), \bfs{y}) = \frac{1}{n(n-1)}\sum_{i=1}^{n}\sum_{j \neq i}\Loss(f(\bfs{x}_{[j,\cdot]}), {y}_i).$$ 
In practice, we usually permute the features of interest multiple times to achieve a similar goal \citep{datta2016algorithmic, fisher2019all}. To study the idea of ``perturbating input'', we first perturb the input vector $\bfs{x}_{[i,\cdot]} = (x_{[i,1]}, x_{[i,2]}, \cdots, x_{[i,p]})^T \in \bfs{X}$, denoted as:  
$$\hat{\bfs{x}}_{[i,\cdot]} = \bfs{\tau} \star \bfs{x}_{[i,\cdot]} = (\tau_1, \tau_2, \cdots, \tau_p)^T \star (x_{[i,1]}, x_{[i,2]}, \cdots, x_{[i,p]})^T , $$
where $\bfs{\tau} = (\tau_1, \tau_2, \cdots, \tau_p)^T$ is the perturbating parameter and the binary operation $\star$ is a component-wise multiplication. We omit it when there is no confusion. We further generalize the representation of features with perturbation as:  
\begin{align} \label{eq:oper}
    \hat{\bfs{z}}_{[\cdot,i]} :&= \hat{\bfs{x}}_{[\cdot,i]} + \bfs{\zeta}_i =(\tau_s x_{[1,i]}, \tau_s x_{[2,i]}, \cdots, \tau_s x_{[n,i]})^T + (\zeta_1, \zeta_2, \cdots, \zeta_n)^T.
\end{align} 

\begin{remark}\label{prop:general_score_fn}
    A general score function $s_{\I}(\bfs{X}, \bfs{y})$ for any order attribution can be estimated by calculating the $\varphi_{i}(\bfs{X}, \bfs{y})$ and $\varphi_{\I}(\bfs{X}, \bfs{y})$. For example, we have $\varphi_{i}(\bfs{X}, \bfs{y}) = \Loss_{emp}(f(\bfs{Z}_{i}), \bfs{y}) - \Loss_{emp}(f(\bfs{X}), \bfs{y})$ for feature importance, where $\bfs{Z}_{i} = [\bfs{x}_{[\cdot, 1]},  \hat{\bfs{z}}_{[\cdot, i]}, ..., \bfs{x}_{[\cdot, p]}]$. Similarly, we can derive $\bfs{Z}_{\I} = [\bfs{x}_{[\cdot, 1]},  \hat{\bfs{z}}_{[\cdot, i]}, ..., \hat{\bfs{z}}_{[\cdot, j]}, \bfs{x}_{[\cdot, p]}]$ for $\I = \{i, j\}$ for higher order attributions.
\end{remark}

\subsubsection{Statistical quantification of axioms}

The statistics from the feature attribution set can be utilized to quantify the sampled Rashomon set according to the fundamental axioms identified above. 

\begin{definition}[Searching efficiency ratio]
    Searching efficiency ratio (SER) is defined as the ratio of the number of valid models in the sampled Rashomon set to the total number of searched models, given by $\text{SER} =|\hat{\R}|/N$, where $N$ is the number of all searched models and $|\hat{\R}|$ denotes the number of valid models. 
\end{definition}
This ratio can be used to test if a searching algorithm satisfies the search efficiency. Normally, a top-down searching algorithm can guarantee the search efficiency property in its model class, while random search may find models outside the Rashomon set.

\begin{definition}[Functional sparsity distance]
    The Chebyshev distance is applied to measure the distance between two feature attribution sets and therefore implies the redundancy of the sampled Rashomon set. The distance is defined as: $$ \forall_{f_{i}, f_{j} \in \R}dist_{\infty}(\bfs{q}_{f_{i}}, \bfs{q}_{f_{j}}) = \lim_{p \to \infty}(\sum_{\I \in \mathbb{I}} \left | q_{\I}(f_{i}) - q_{\I}(f_{j}) \right |)^{(1/p)}.$$ 
\end{definition}
Ideally, the distance between two feature attribution sets, equivalently, any two sampled models, should not be 0, so that no model in the sampled Rashomon set is functionally redundant.

 \begin{definition}[Functional efficiency range]
     Functional efficiency range (FER) is the sum of the maximum importance range. The interval of a specific feature attribution within a sampled Rashomon set $[\min(\bfs{q}_{\I}), \max(\bfs{q}_{\I})]$ can be seen as the range of a feature attribution across the sampled Rashomon set \citep{fisher2019all, li2023exploring, hsu2022rashomon}. The individual feature attribution can be calculated as $\max(\bfs{q}_{\I}) - \min(\bfs{q}_{\I})$ and total range is defined as: $$\text{FER}_{\epsilon} = \sum_{\I \in \mathbb{I}}\max(\bfs{q}_{\I}) - \min(\bfs{q}_{\I}).$$
 \end{definition}
This metric facilitates a direct comparison of feature attribution ranges across various sampling methods.


\subsection{Constrained Non-Differentiable Optimization}
The connection between the Proposition \ref{prop:general_rs} and \ref{prop:general_score_fn} builds the basis for our sampling method. The problem becomes perturbing the feature attribution space on the reference model within the Rashomon set. According to the axiom functional sparsity and functional efficiency ratio, we define a constrained non-differentiable optimization problem as: 
\begin{align}
\left.\begin{matrix}
\inf_{\bfs{\tau} \in \mathbb{R}^{d}, \bfs{\zeta} \in \mathbb{R}^{n}} q_{\I}(f)
\\ 
\sup_{\bfs{\tau} \in \mathbb{R}^{d}, \bfs{\zeta} \in \mathbb{R}^{n}} q_{\I}(f)
\end{matrix}\right\}\text{s.t. } f \in \R \text{ and } & \forall_{f_{i}, f_{j} \in \R} dist_{\infty}(\bfs{q}_{f_{i}}, \bfs{q}_{f_{j}}) \neq 0.  
\end{align}

\begin{lemma}[Theoretical bounds analysis without constraints]
    In theory, any attribution assigned by the generalized feature attribution function is greater than or equal to 0, from Definition \ref{def:fat}.
    $$\bfs{Q}_{\mathbb{I}}(\R) \subseteq \mathbb{R}^{+}_{0}.$$
\end{lemma}
\noindent
\textbf{Proof. } With the assumption that the reference model is near optimal, any perturbation in a feature increases error\footnote{In the case of loss decreasing, the fact points out a requirement for further optimization.}, or the error remains the same, indicating that the feature is redundant. 
In terms of the absolute interaction strength, the value of zero means that no interaction happens among features and that is the minimum strength. Negative values indicate effect direction. 
\begin{lemma}[Theoretical bounds analysis with Rashomon constraints]\label{lemma:grs}
By Proposition \ref{prop:general_rs}, the Rashomon set is equivalently defined using $\epsilon$.
The Rashomon set definition can be further rewritten as:
\begin{equation}
    0 \leq \sup_{(f_{ref} \circ m) \in \R} \Loss_{ref}(f_{ref}(\bfs{Z}, \bfs{y})) - \Loss_{ref}(f_{ref}(\bfs{X}, \bfs{y})) \leq \epsilon,
\end{equation}
where $\bfs{X} = \inf_{Z\in \mathbb{R}^{n \times p}} \Loss_{ref}(f_{ref}(\bfs{Z}))$ and $f_{ref}(\bfs{Z}) \leftrightarrow f_{ref} \circ m(\bfs{X})$.
\end{lemma}

\begin{theorem}\label{theory:e_sub}
    For a vector $\bfs{Z}$ that makes $f_{ref}(\cdot)$ within the Rashomon set and $\epsilon \geq 0$, we can further derive that \begin{equation}\label{eq:e_sub}\forall_{(f_{ref} \circ m) \in \R} \Loss(f_{ref}(\bfs{Z})) \leq \Loss(f_{ref}(\bfs{X})) + \left \langle \bfs{Z}-\bfs{X}, \bfs{X}' \right \rangle + \epsilon,
    \end{equation} from Lemma \ref{lemma:grs}. 
    The maximization problem then can be converted into a minimization problem by negating the objective function:
    $\forall_{(f_{ref} \circ m) \in \R} \Loss(f_{ref}(\bfs{Z})) \geq \Loss(f_{ref}(\bfs{X})) + \left \langle \bfs{Z}-\bfs{X}, \bfs{X}' \right \rangle  - \epsilon$.
\end{theorem}
\noindent
\textbf{Proof.} We can observe that our problem in Eq. \eqref{eq:e_sub} can be seen as the definition of a concave function. By negating the objective function, it would be definition of the $\epsilon$-subgradient in \citep{bertsekas1973descent, bertsekas2003convex, shor2012minimization}, 
where $\bfs{X}'$ is a subgradient of the complex function $\Loss(f_{ref}(\cdot))$ at $\bfs{X}$, which does not require the differentiability of the function, and $\left \langle \cdot, \cdot \right \rangle$ denotes the usual inner product. The set $\partial_{\epsilon}\Loss(f_{ref}(\bfs{X}))$ of all subgradients at $\bfs{X}$ will be called subdifferential of $\Loss(f_{ref}(\cdot))$ at $\bfs{X}$.


\begin{theorem}\label{theory:13}
    The subgradient set $\partial_{\epsilon}\Loss(f_{ref}(\bfs{X}))$ characterizes the sampled Rashomon set by Theorem \ref{theory:e_sub} and it is evident that we have $$ 0 \leq \epsilon_{1} \leq \epsilon_{2} \leftrightarrow \partial \Loss(f_{ref}(\bfs{X})) \subseteq \partial_{\epsilon_{1}}\Loss(f_{ref}(\bfs{X})) \subseteq \partial_{\epsilon_{2}}\Loss(f_{ref}(\bfs{X})),$$ which corresponds to the Rashomon set property. 
\end{theorem}
\noindent

\begin{proposition}
    The set $\partial_{\epsilon} \Loss(f_{ref}(\bfs{X}))$ is characterized by the following property, which is adapted from \cite{rockafellar1997convex}.
\begin{equation}
    \partial_{\epsilon} \Loss(f_{ref}(\bfs{X})) = \{ \bfs{X}' \mid \left \langle \bfs{X}, \bfs{X}' \right \rangle - \Loss^{*}(f_{ref}(\bfs{X}')) - \Loss(f_{ref}(\bfs{X}))  \leq \epsilon \},
\end{equation}
where $$\Loss^{*}(f_{ref}(\bfs{X}')) = \sup_{\bfs{X}}\{\Loss(f_{ref}(\bfs{X})) - \left \langle \bfs{X}, \bfs{X}' \right \rangle \},$$ is the conjugate concave function of $\Loss$. The support function of $\partial_{\epsilon} \Loss(f_{ref}(\bfs{X}))$ is given by the useful equation in \citep{bertsekas1973descent}.
\begin{equation}\label{eq:supporting_fn}
    \sigma [\bfs{z}|\partial_{\epsilon} \Loss(f_{ref}(\bfs{X}))] =
     \sup_{\bfs{X}' \in \partial_{\epsilon} \Loss(f_{ref}(\bfs{X}))}\left \langle \bfs{z}, \bfs{X}' \right \rangle = \inf_{\lambda>0}\frac{\Loss(f_{ref}(\bfs{X})) - \Loss(f_{ref}(\bfs{X}+ \lambda \bfs{z})) + \epsilon}{\lambda}.
\end{equation}
\end{proposition}

\begin{remark}
    It is noted that the subgradient method is not a descent method and is usually to keep track of the point with the smallest function value.
    This property benefits the Rashomon set sampling, as our goal is to find a set of well-performing models.
\end{remark}

\begin{proposition}\label{prop:rset_condition}
    Let $\bfs{X}$ be a vector such that $\Loss(f_{ref}(\bfs{X})) \geq 0$. Then
    $$0 \in \partial_{\epsilon}\Loss(f_{ref}(\bfs{X}))
    \leftrightarrow
     0 \leq \Loss(f_{ref}(\bfs{X})) - \Loss(f_{ref}(\bfs{X}^{*})) \leq \epsilon. $$ 
\end{proposition}
\textbf{Proof. } As the $\Loss$ is near-optimal at $\bfs{X}^{*}$, thus there always exists $0 \in \partial_{\epsilon}\Loss(f_{ref}(\bfs{X}^{*}))$. By definition in Theorem \ref{theory:e_sub}, we can obtain $0 \in \partial_{\epsilon}\Loss(f_{ref}(\bfs{X})) \to \Loss(f_{ref}(\bfs{X})) \leq \Loss(f_{ref}(\bfs{X}^{*})) + \epsilon$. The property ensures the Rashomon set condition from the sampling process.

\begin{proposition}\label{prop:init_sampling}
    Let $\bfs{X}^{*} = \inf_{\bfs{X}} \Loss_{\epsilon}(f_{ref}(\bfs{X}))$ be the initial sampling vector such that $0 \leq \Loss(f_{ref}(\bfs{X}^{*})) < \infty$. 
    Then, for any vector $\bfs{z}$, we have 
    $$\sup_{\bfs{X}' \in \partial_{\epsilon} \Loss(f_{ref}(\bfs{X}^{*}))}\left \langle \bfs{z}, \bfs{X}' \right \rangle \geq 0.$$
\end{proposition}
\textbf{Proof. } Given the 
$0 \in \partial_{\epsilon}\Loss(f_{ref}(\bfs{X}^{*}))$, we can use the supporting function \eqref{eq:supporting_fn} and derive
$$\inf_{\lambda>0}\frac{\Loss(f_{ref}(\bfs{X}^{*})) - \Loss(\bfs{X}^{*} + \lambda \bfs{z})}{\lambda} = 0.$$ 
 Intuitively, $\epsilon \geq 0$, and
\begin{equation}
\inf_{\lambda>0}\frac{\Loss(f_{ref}(\bfs{X}^{*})) - \Loss(\bfs{X}^{*} + \lambda \bfs{z}) + \epsilon}{\lambda} = \sup_{\bfs{X}' \in \partial_{\epsilon} \Loss(f_{ref}(\bfs{X}^{*}))}\left \langle \bfs{z}, \bfs{X}' \right \rangle \geq 0.
\end{equation}
\begin{remark}
    This property ensures that any direction of ascent $\bfs{z}$ can be searched from the initial sampling point.
\end{remark}

\begin{proposition}
    Similarly, let $\bfs{z}$ be a vector such that $$\sup_{\bfs{X}' \in \partial_{\epsilon} \Loss(f_{ref}(\bfs{X}))}\left \langle \bfs{z}, \bfs{X}' \right \rangle < 0,$$ then there holds
    $$\Loss(f_{ref}(\bfs{X})) - \inf_{\lambda>0}\Loss(\bfs{X} + \lambda \bfs{z})  + \epsilon < 0.$$
\end{proposition}
\textbf{Proof. }This can be proved by using the supporting function. This property states that the value of $\Loss(\bfs{X} + \lambda \bfs{z})$ increases by an ascent along a direction $\bfs{z}$, and it will exceed the value $\Loss(f_{ref}(\bfs{X}))$ by more than $\epsilon$.

\begin{theorem}
    The Rashomon set can be sampled by gradually increasing $\lim_{\epsilon_{i} \to \hat{\epsilon}}$, where $\hat{\epsilon}$ is the pre-defined tolerance in the Rashomon set, and recording the searched point. 
\end{theorem}
\textbf{Proof. }As stated in Theorem \ref{theory:13}, we can include subgradient sets by $$\lim_{\epsilon_{i} \to \hat{\epsilon}}\partial \Loss_{\hat{\epsilon}}(\bfs{X}) = \partial\Loss(f_{ref}(\bfs{X})) \cup \partial_{\epsilon_{1}}\Loss(f_{ref}(\bfs{X})) \cup \ldots \cup \partial_{\epsilon_{i}}\Loss(f_{ref}(\bfs{X})),$$
where each subgradient set corresponds to a set of models within the Rashomon set. Thus we can have:
$$\hat{\R}(\hat{\epsilon}, f_{ref}, \F) = \{f_{ref}\} \cup \{f_{1}(\bfs{X}), f_{2}(\bfs{X}) \} \cup \ldots \cup \{ f_{1}(\bfs{X}) \ldots f_{i}(\bfs{X})\}.$$

\subsubsection{Sampling algorithm steps}
The above theorems and propositions form the basis for the sampling algorithm.
\begin{enumerate}
    \item Given the pre-defined $\hat{\epsilon}$ and initial sampling vector $\bfs{X}^{*} = \inf_{X}\Loss_{ref}(\bfs{X})$, we set $\epsilon_{0} = 0$ and $\epsilon_{i} = \gamma \epsilon_{i-1}$, where $\gamma$ controls the $\epsilon$ update. $$\hat{\R}(\epsilon_{0}, f_{ref}, \F) = \{f_{ref}\}.$$
    \item Find an ascent direction $\bfs{z}$, such that $$\sup_{\bfs{X}' \in \partial_{\epsilon_{0}} \Loss(f_{ref}(\bfs{X}^{*}))}\left \langle \bfs{z}, \bfs{X}' \right \rangle \geq 0.$$
    Based on Proposition \ref{prop:init_sampling}, we set the direction according to the features $(\bfs{x}_{[\cdot,i]})_{i=1}^{d}$ for the purpose of feature attribution giving a set of directions $(\bfs{z}_{i})_{i=1}^{d}$, which enables us to calculate the main effects $\bfs{\varphi}$ for features. Higher-order feature attributions can be derived similarly, as: 
    $$ \sup_{\bfs{X}' \in \partial_{\epsilon_{0}} \Loss(f_{ref}(\bfs{X}^{*}))}\left \langle \bfs{z}_{i}, \bfs{X}' \right \rangle \geq 0, \text{ for all } \bfs{z}_{i} \text{ in } (\bfs{z}_{i})_{i=1}^{d}.$$
    \item For each direction $\bfs{z}_{i}$, find a step size $\lambda_{k}$ such that 
    $$\Loss(f_{ref}(\bfs{X})) - \inf_{\lambda>0}\Loss(\bfs{X} + \lambda_{k} \bfs{z}_{i})  + \epsilon_{i} < 0$$
    \item Set $\bfs{X^{(k+1)}} = \bfs{X^{(k)}} + \lambda_{k} \bfs{z}_{i}$, where $\lambda_{k} > 0$ and $$ \epsilon_{i} \leq \Loss_{ref}(\bfs{X}^{(k+1)})-\Loss_{ref}(\bfs{X}^{k}) \leq  \hat{\epsilon},$$

    At each step, we can find the subgradient set
    $$\partial \Loss_{\epsilon_{i-1}}(\bfs{X}) = \partial\Loss(f_{ref}(\bfs{X})) \cup \partial_{\epsilon_{1}}\Loss(f_{ref}(\bfs{X})) \cup \ldots \cup \partial_{\epsilon_{i-1}}\Loss(f_{ref}(\bfs{X})).$$

    \item We update the $\epsilon_{i} = \gamma \epsilon_{i-1}$ and return to step 3 until $\epsilon_{i} \to \hat{\epsilon}$ $$\hat{\R}(\epsilon_{i}, f_{ref}, \F)= \{f_{ref}(\bfs{X^{(1)}}), f_{ref}(\bfs{X^{(2)}}), f_{ref}(\bfs{X^{(3)}}) \ldots f_{ref}(\bfs{X^{(k)}})\}.$$
\end{enumerate}







\section{Practical Axioms in Our Framework}\label{sec:practical-axioms}

In this section, we illustrate why our method satisfies all axioms proposed above. The model structure generalizability and model evaluation generalizability are guaranteed by the generalized Rashomon set from Proposition \ref{prop:general_rs} and Definition \ref{def:grs}. The SER in searching algorithm remains 1, $\text{SER} =|\hat{\R}|/N = 1$, ensured by Proposition \ref{prop:rset_condition}.

\subsection{Functional Sparsity}
To satisfy functional sparsity, we want to show $dist_{\infty}(\bfs{q}_{f_{i}}, \bfs{q}_{f_{j}}) > 0$ for all $f_{i}, f_{j} \in \R$. 
This can be proved as follows: for any two models in the Rashomon set, there are two corresponding feature attribution sets. As long as one of the feature attributions in the set is different, then their distance is greater than 0. 

\begin{theorem}
    For any two different models in our sampled Rashomon set, we have at least one different feature attribution
    \begin{equation}
        \forall_{f_{i}, f_{j} \in \hat{\R}(\hat{\epsilon}, f_{ref}, \F)}dist_{\infty}(\bfs{q}_{f_{i}}, \bfs{q}_{f_{j}}) > 0.
    \end{equation}
\end{theorem}
\noindent
\textbf{Proof. }From step 2 in our algorithm, we know that the sampling direction is along with features. For any two sampled models, there are two scenarios: (1) two models sampled along with different features and (2) two models sampled along with the same features. Here we will discuss both using $s$ and $t$ as feature indices. 

In case $s \neq t$, we assume that model $f_{i}$ is sampled along $\bfs{z}_{s}$, while model $f_{j}$ is sampled along $\bfs{z}_{t}$. Our target is to prove $\left | q_{s}(f_{i}) - q_{s}(f_{j}) \right | \neq 0$ or $\left | q_{t}(f_{i}) - q_{t}(f_{j}) \right | \neq 0$. Based on the general feature attribution function and Proposition \ref{prop:general_score_fn}, we can approximate the feature attribution by\footnote{We use superscript to distinguish models. }
\begin{align}
    \varphi_{s}^{j}(\bfs{X}, \bfs{y})  
    & = \E[\Loss(f_{i}(\bfs{X}_{\setminus s}), \bfs{y})] - \E[\Loss(f_{i}(\bfs{X}), \bfs{y})] \\ \nonumber
    & \approx \Loss_{emp}(f_{i}(\hat{\bfs{Z}}_{\setminus s}), \bfs{y}) - \Loss_{emp}(f_{i}(\hat{\bfs{Z}}), \bfs{y}) \\ \nonumber
    & \approx  \Loss_{emp}(f_{i}(\tau_{s}\bfs{X} + \bfs{\zeta}_{s}), \bfs{y}) - \Loss_{emp}(f_{i}(\tau_{s}\bfs{X}), \bfs{y}).
\end{align}

Given the model generalization representation Eq. \eqref{eq:grs_tau} in Proposition \ref{prop:general_rs} $\Loss_{emp}(f_{i}(\bfs{X})) = \Loss_{emp}(f_{ref}(\bfs{X} + \lambda_{k}\bfs{z}_{s}))$  we have
\begin{align}
    \Loss_{emp}(f_{i}(\tau_{s}\bfs{X} + \bfs{\zeta}_{s}), \bfs{y}) & = \Loss_{emp}(f_{ref}(\tau_{s}\bfs{X} + \bfs{\zeta}_{s} + \lambda_{k}\bfs{z}_{s}), \bfs{y}) \\ \nonumber
    & = \Loss_{emp}(f_{ref}(\hat{\tau}_{s}\bfs{X} + \bfs{\zeta}_{s}), \bfs{y}).
\end{align}
where $\hat{\tau}_{s}\bfs{x}_{s} = \tau_{s}\bfs{x}_{s} + \lambda_{k}\bfs{z}_{s}$.  Similarly, we can derive that $\Loss_{emp}(f_{i}(\tau_{s}\bfs{X}), \bfs{y}) = \Loss_{emp}(f_{ref}(\hat{\tau}_{s}\bfs{X}), \bfs{y})$. We therefore approximate $ \varphi_{s}^{i}(\bfs{X}, \bfs{y})$ by
\begin{align}
    \left |\varphi_{s}^{i}(\bfs{X}, \bfs{y})\right | \approx  \left | \left \langle (\hat{\tau}^{i}_{s}\bfs{X} + \bfs{\zeta}_{s} - \hat{\tau}^{i}_{s}\bfs{X}), (\hat{\tau}^{i}_{s}\bfs{X})'\right \rangle  \right | \approx  \left | \left \langle  \bfs{\zeta}_{s}, (\hat{\tau}^{i}_{s}\bfs{X})'\right \rangle \right |.
\end{align}

\noindent For model $f_{j}$, the attribution of feature can be derived as:
\begin{align}
    \Loss_{emp}(f_{j}(\tau_{s}\bfs{X} + \bfs{\zeta}_{s}), \bfs{y}) & = \Loss_{emp}(f_{ref}(\tau_{s}\bfs{X} + \bfs{\zeta}_{s} + \lambda_{k}\bfs{z}_{t}), \bfs{y}) \\ \nonumber
    & = \Loss_{emp}(f_{ref}(\tilde{\bfs{\tau}}_{s;t}\bfs{X} + \bfs{\zeta}_{s}), \bfs{y}),
\end{align}
where $\hat{\bfs{x}}_{[\cdot,t]} = \bfs{x}_{[\cdot,t]} + \lambda_{k}\bfs{z}_{t}$ and $\hat{\bfs{x}}_{[\cdot,s]} = \tau_{s}\bfs{x}_{[\cdot,s]}$ remains same. We can obtain $ \left |\varphi_{s}^{j}(\bfs{X}, \bfs{y})\right | \approx  \left | \left \langle  \bfs{\zeta}_{s}, (\tilde{\bfs{\tau}}_{s;t}\bfs{X})'\right \rangle \right |.$ The distance between any two models on different feature attributions can be simplified based on the scaling property of subgradients
\begin{equation}\label{eq:approx_phi}
    \left | \varphi_{s}^{i}(\bfs{X}, \bfs{y}) - \varphi_{s}^{j}(\bfs{X}, \bfs{y}) \right | \approx  \left |\left \langle  \bfs{\zeta}_{s}, (\hat{\tau}^{i}_{s}\bfs{X})'\right \rangle -  \left \langle  \bfs{\zeta}_{s}, (\tilde{\bfs{\tau}}_{s;t}\bfs{X})'\right \rangle \right | \approx \left | \left \langle  \bfs{\zeta}_{s}, (\tilde{\bfs{\tau}}_{s;t} - \hat{\tau}^{i}_{s}) \bfs{X}'\right \rangle \right |.
\end{equation}

\noindent From the results, we can observe that the feature attribution difference can only be 0 if $\bfs{\zeta}_{s} = \left |\bfs{0}\right |$, which only happens if such a feature is a constant.

In case $s = t$, we can derive an equation similar  to Eq. \eqref{eq:approx_phi}, such as:
\begin{equation}\label{eq:approx_phi_same_model}
    \left | \varphi_{s}^{i}(\bfs{X}, \bfs{y}) - \varphi_{s}^{j}(\bfs{X}, \bfs{y}) \right | \approx  \left |\left \langle  \bfs{\zeta}_{s}, (\hat{\tau}^{i}_{s}\bfs{X})'\right \rangle -  \left \langle  \bfs{\zeta}_{s}, (\hat{\tau}^{j}_{s}\bfs{X})'\right \rangle \right | \approx \left | \left \langle  \bfs{\zeta}_{s}, (\hat{\tau}^{i}_{s} - \hat{\tau}^{j}_{s}) \bfs{X}'\right \rangle \right |, 
\end{equation}
\noindent where $\hat{\tau}^{i}_{s} \neq \hat{\tau}^{j}_{s}$, as two models are different. The proof is complete.

\subsection{Functional Efficiency Analysis}
\begin{theorem}
    The extreme feature attribution values are obtained when the model is sampled along with the direction of such a feature, e.g., $\bfs{z}_{s}$.
\end{theorem}
\textbf{Proof. } Considering the feature attribution set $\bfs{q}_{\mathbb{I}}(f_{ref})$ as a baseline, the attribution of a specific feature, 
e.g., $s$, reaches its maximum value when the difference between the current model and the reference model is at its peak, i.e., $\max (q_{s}(f_{i}) - q_{s}(f_{ref}))$. Conversely, the minimum occurs at $\min (q_{s}(f_{i}) - q_{s}(f_{ref}))$. 

We assume that the model $f_{i}$ is not sampled along with the direction of feature $s$, so we know that the difference depends on the $(\tilde{\bfs{\tau}}^{i}_{s;t} - \hat{\tau}^{ref}_{s})$ from Eg. \eqref{eq:approx_phi}. Such a difference only changes when the model $f_{i}$ is sampled along with the direction of the feature $\bfs{z}_{s}$, which contradicts our assumption.

To calculate the maximum importance range, we apply the Eq. \eqref{eq:approx_phi_same_model} to a specific feature and have 
\begin{equation}
    \max (q_{s}(f_{i}) - \min (q_{s}(f_{j})) = \left \langle  \bfs{\zeta}_{s}, \sup_{\hat{\tau}^{i}_{s}}\hat{\tau}^{i}_{s}\bfs{X}' - \inf_{\hat{\tau}^{j}_{s}}\hat{\tau}^{j}_{s} \bfs{X}'\right \rangle, 
\end{equation}
\noindent where the infimum and supremum can be found from the recorded subgradient set. The number of sampled models depends on the pre-defined $\epsilon$ and update rate. We will show their comparison in experiments.



\section{Results and Discussion}\label{sec:experiments}

In this section, we illustrate the generalizability of our framework by applying it to a synthetic data set and statistically comparing it with different sampling methods in the current literature.

\subsection{Generalizability in Synthetic Data Set}\label{subsec:generalizability-in-syntheticdata}
\subsubsection{Problem setting}
To illustrate the generalizability, we designed an experiment in a synthetic dataset and aimed at solving the quadratic problem. The problem is to train a machine learning model to solve the quadratic equation $ax^2 + bx + c = 0$, where the variables $a$, $b$, and $c$ are inputs and $x_1$ and $x_2$ are the outputs. Based on mathematical principles, two ground truth error-free models are $f_{1}=\frac{-b+\sqrt{b^2-4ac}}{2a}$ and $f_{2}=\frac{-b-\sqrt{b^2-4ac}}{2a}$. The regression problem can be fit using different types of models. Here we applied multi-layer perceptron (MLP)  and tree-based ensemble models, Random Forest (RF), to fit the data set under supervised learning settings and physically informed NNs (PINN) for unsupervised learning. 

\begin{figure}[h!]
     \centering
     \begin{subfigure}{.32\textwidth}
         \centering
         \includegraphics[height=1.4in]{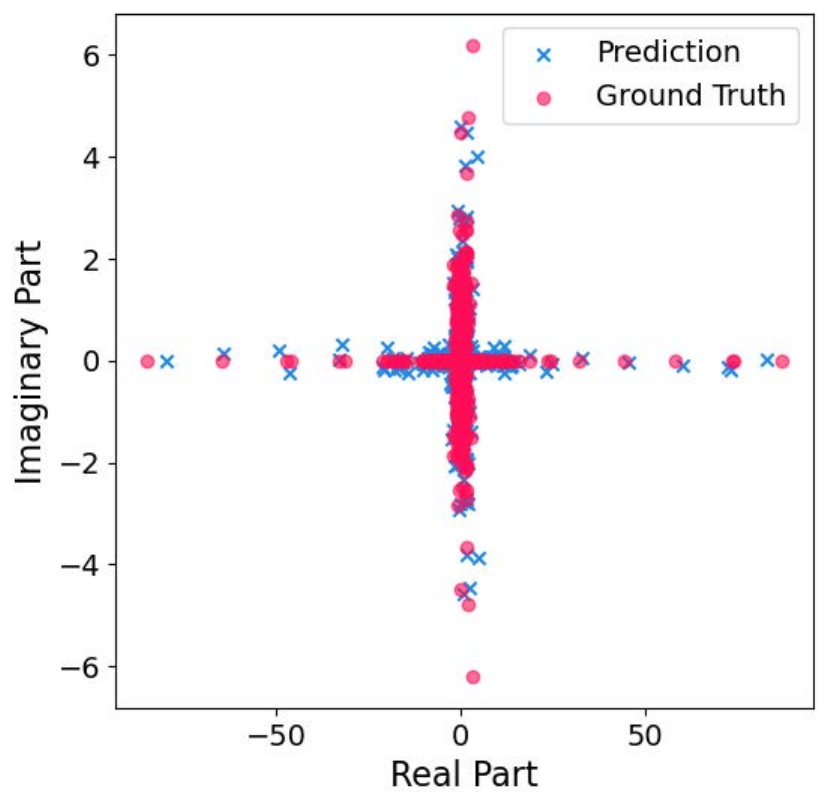}
         \caption{An well-trained MLP with R2 0.99 and MAE 0.03 in testing set}
         \label{fig:three sin x}
     \end{subfigure}
     \hfill
     \begin{subfigure}{.32\textwidth}
         \centering
         \includegraphics[height=1.4in]{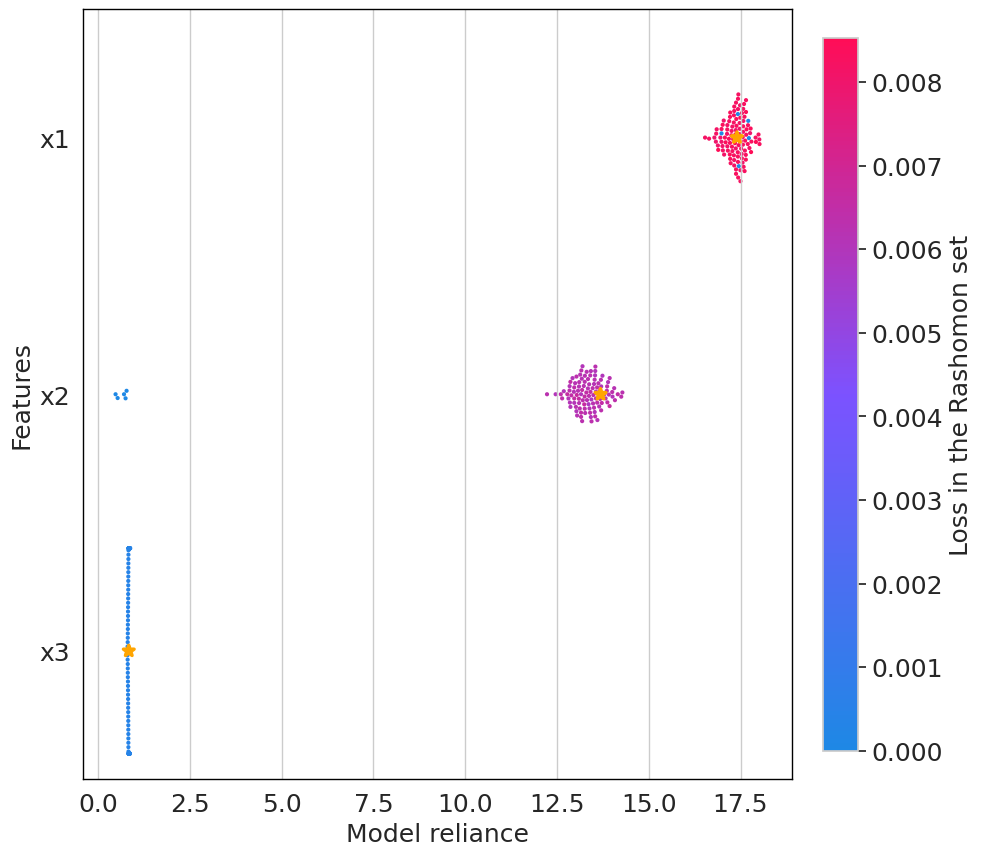}
         \caption{First-order feature attribution based on the MLP.}
         \label{fig:three sin x}
     \end{subfigure}
    \hfill
     \begin{subfigure}{.32\textwidth}
         \centering
         \includegraphics[height=1.4in]{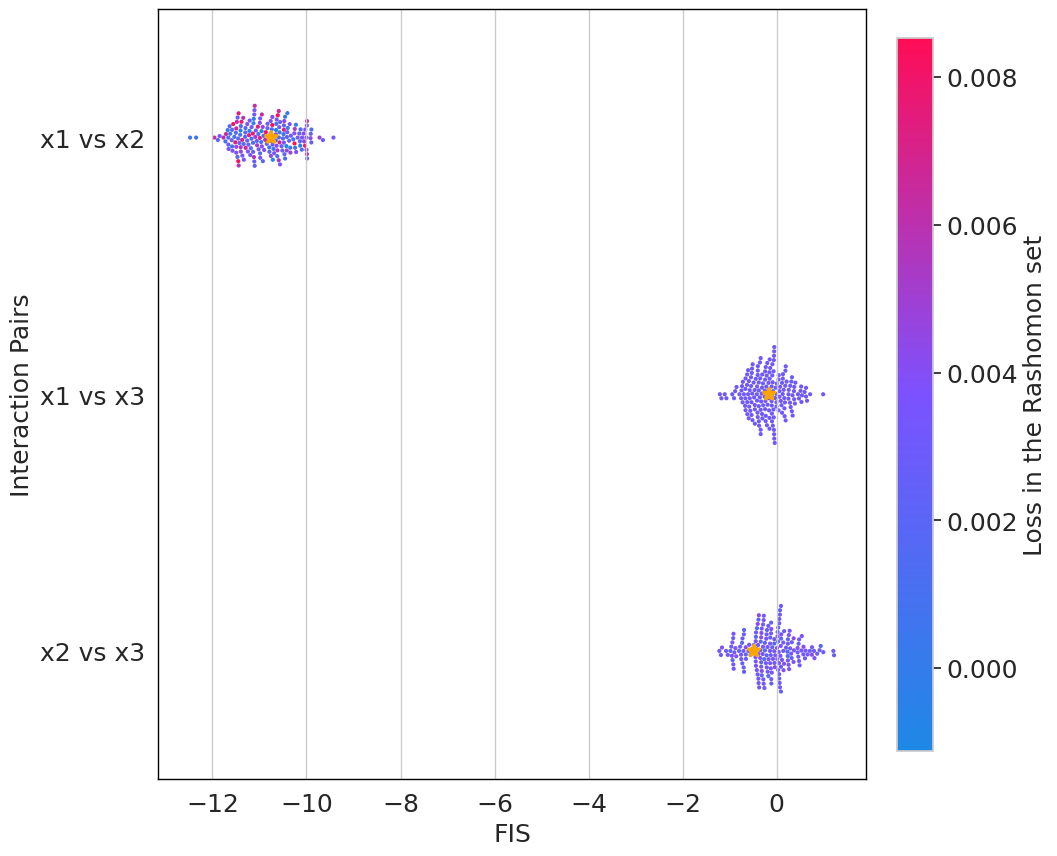}
         \caption{Second-order feature attribution based on the MLP.}
         \label{fig:three sin x}
     \end{subfigure}
     \vfill
     \begin{subfigure}{.32\textwidth}
         \centering
         \includegraphics[height=1.4in]{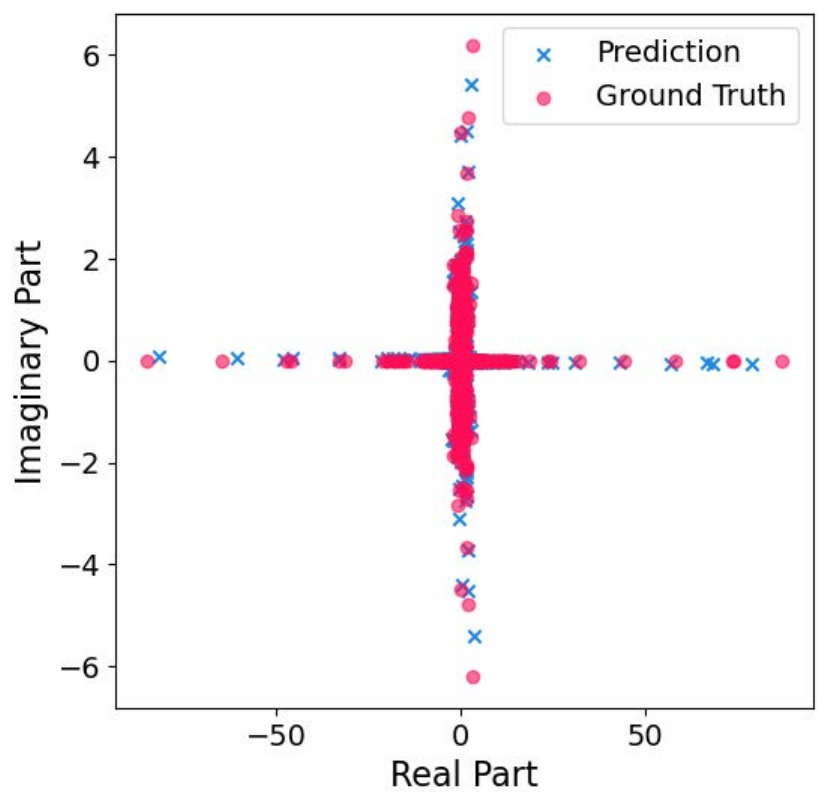}
         \caption{A well-trained PINN with R2 0.99 and MAE 0.03 in testing set}
         \label{fig:three sin x}
     \end{subfigure}
     \hfill
     \begin{subfigure}{.32\textwidth}
         \centering
         \includegraphics[height=1.4in]{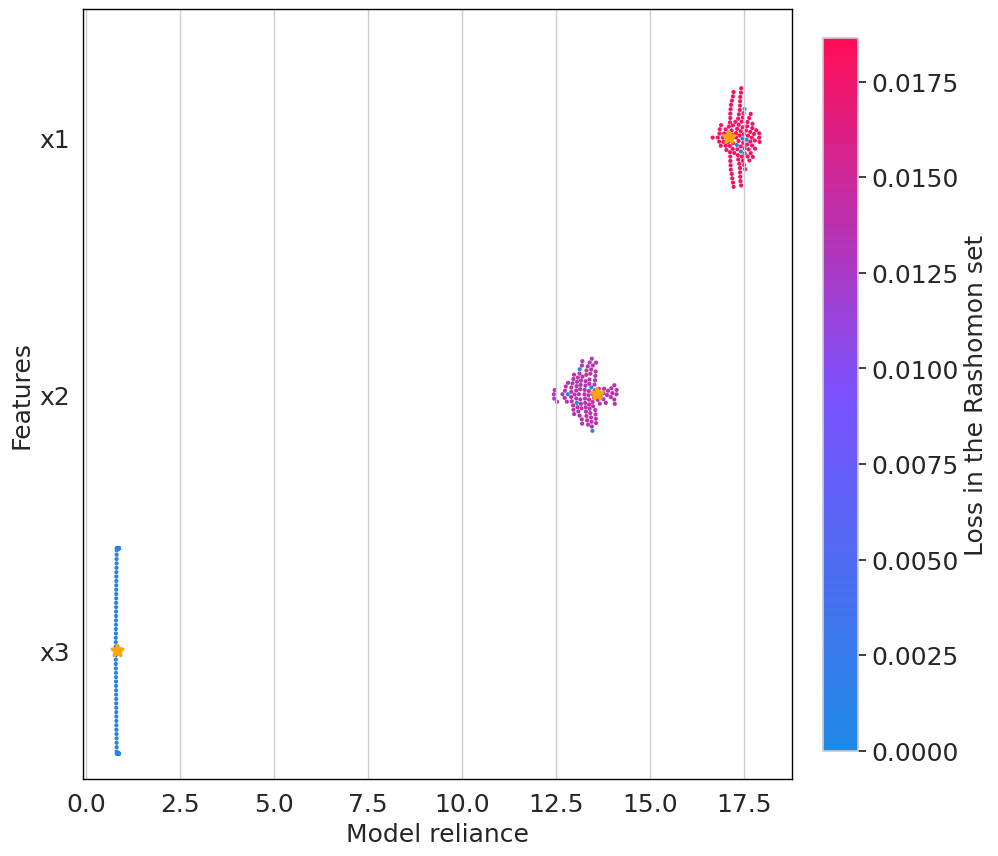}
         \caption{First-order feature attribution based on the PINN.}
         \label{fig:three sin x}
     \end{subfigure}
    \hfill
     \begin{subfigure}{.32\textwidth}
         \centering
         \includegraphics[height=1.4in]{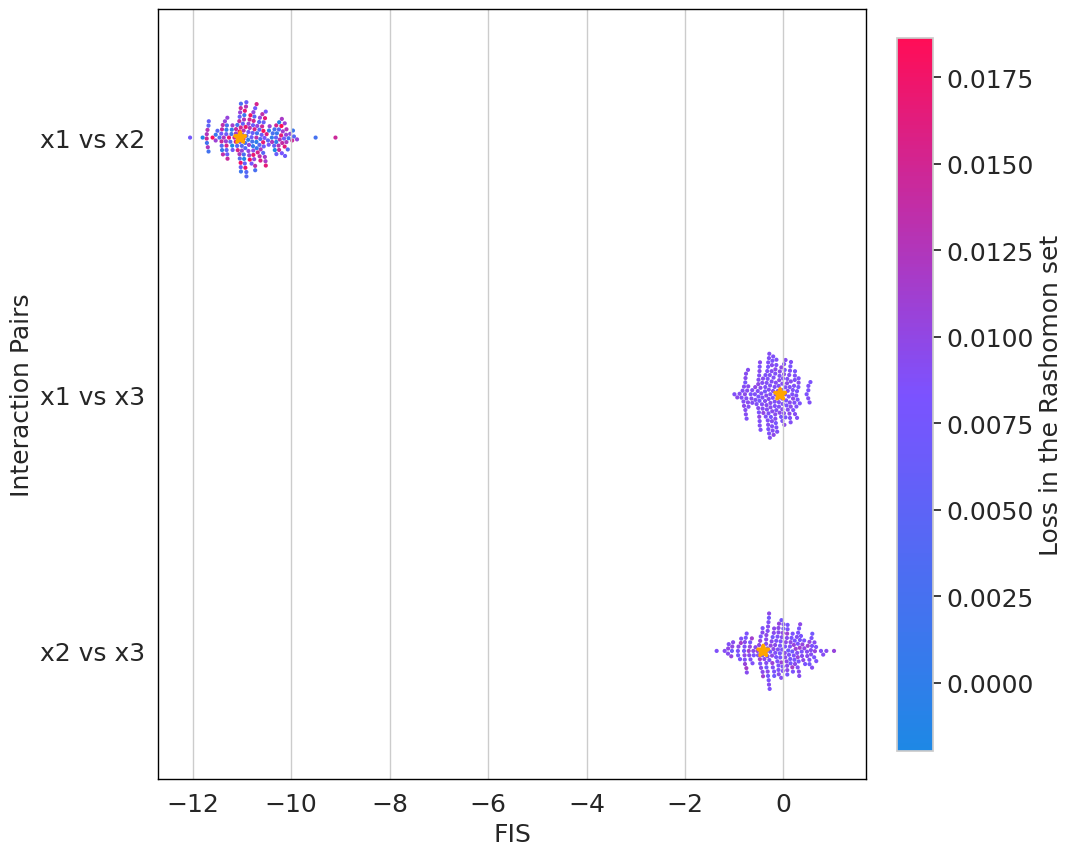}
         \caption{Second-order feature attribution based on the PINN.}
         \label{fig:three sin x}
     \end{subfigure}
     \vfill
     \begin{subfigure}{.32\textwidth}
         \centering
         \includegraphics[height=1.4in]{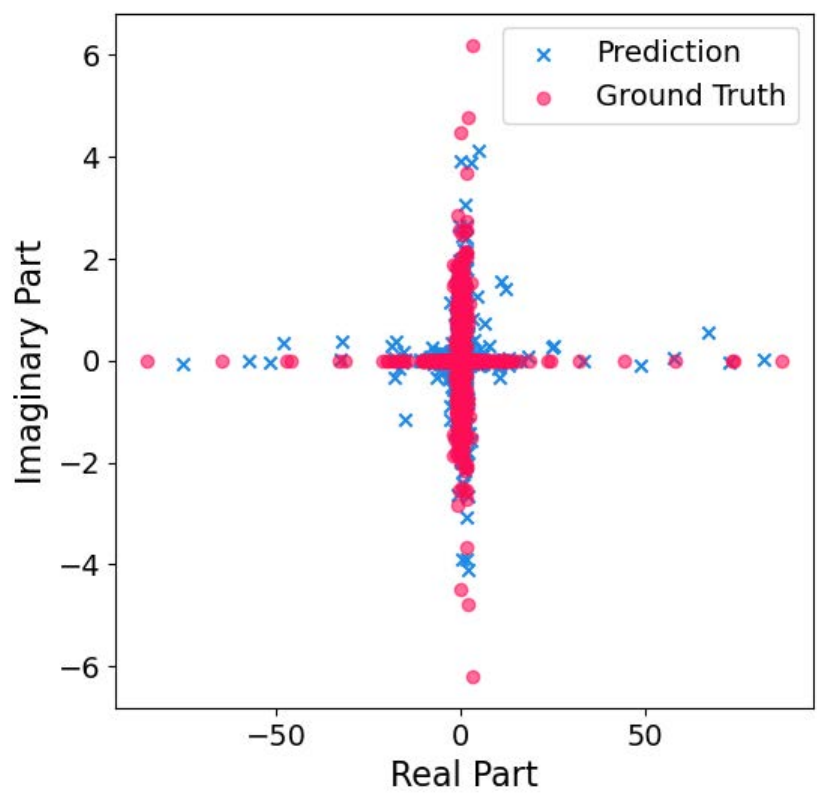}
         \caption{A well-trained RF with R2 0.98 and MAE 0.05 in testing set}
         \label{fig:three sin x}
     \end{subfigure}
     \hfill
     \begin{subfigure}{.32\textwidth}
         \centering
         \includegraphics[height=1.4in]{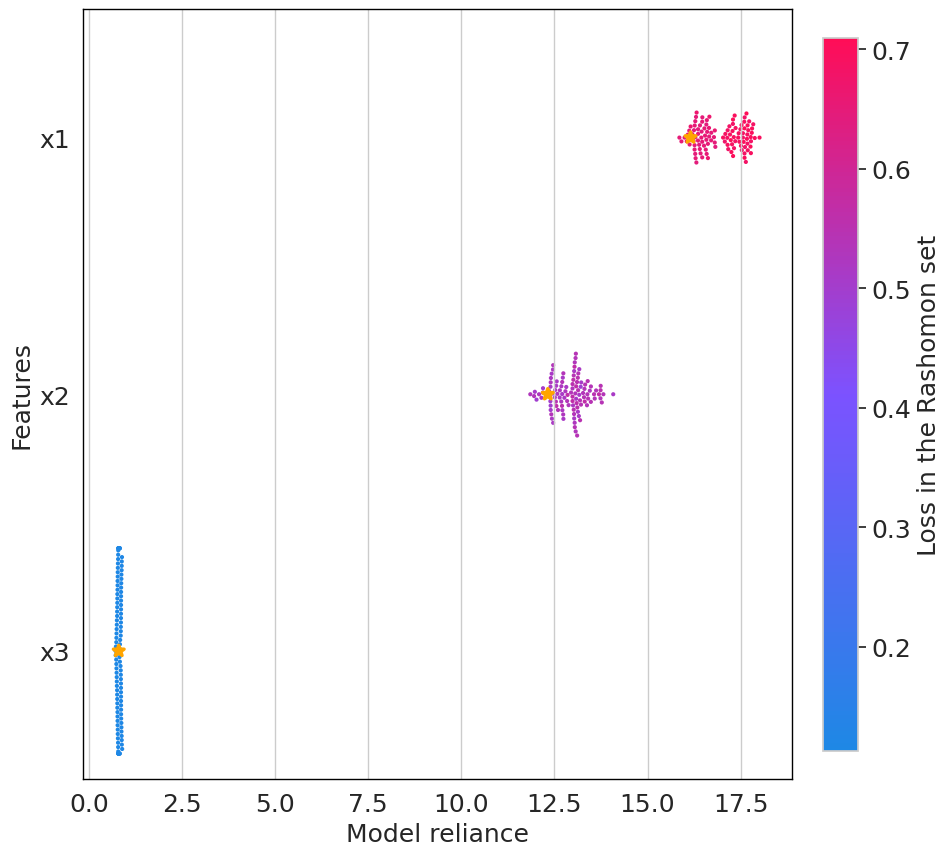}
         \caption{First-order feature attribution based on the RF.}
         \label{fig:three sin x}
     \end{subfigure}
    \hfill
     \begin{subfigure}{.32\textwidth}
         \centering
         \includegraphics[height=1.4in]{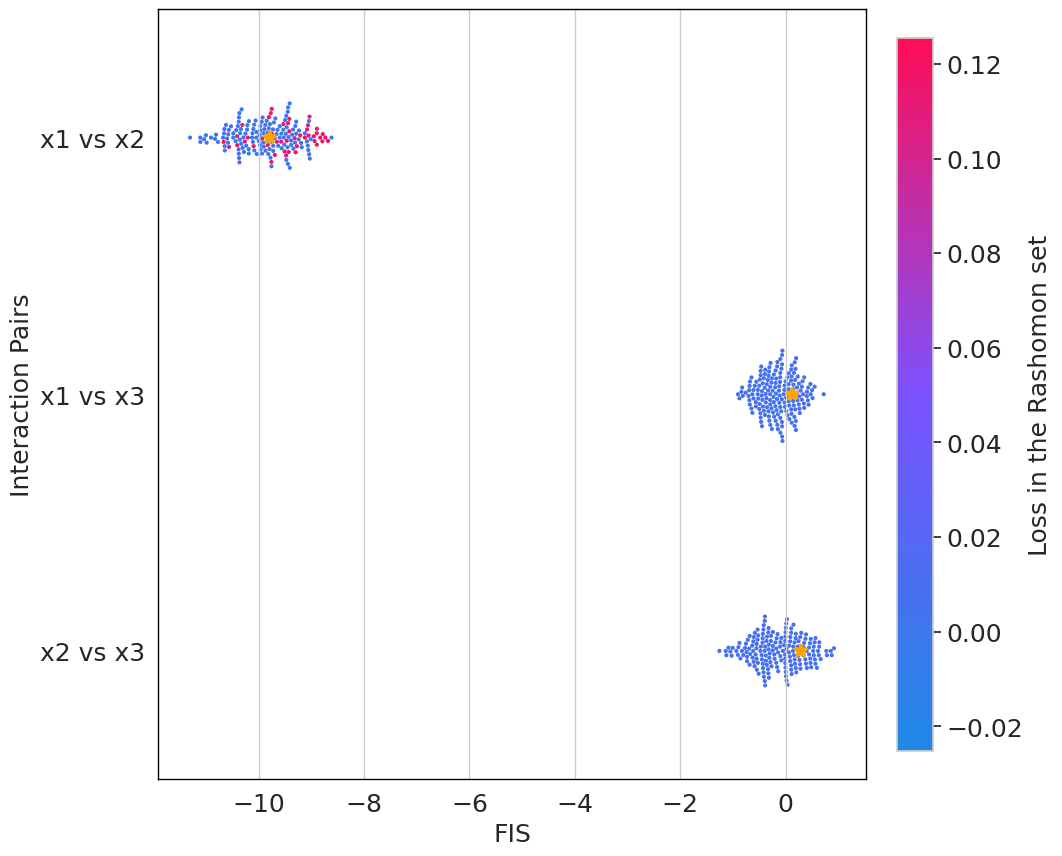}
         \caption{Second-order feature attribution based on the RF.}
         \label{fig:three sin x}
     \end{subfigure}
    \caption{Illustration of model flexibility and explanation flexibility across different learning paradigms—supervised, unsupervised, and ensemble learning—alongside visualizations of first-order and second-order explanation spaces from our proposed framework. The $x$-axis represents the importance value of features, while the $y$-axis denotes corresponding features, with the point color indicating the magnitude of loss (darker shades denote higher loss).} 
\label{fig:illustration of model flexibility and explanation flexibility}
\end{figure}

Given the function, we randomly sampled 12,000 data points from the uniform distribution $a \sim \mathcal{U}(0.01, 1)$\footnote{We set 0.01 instead of 0 for fast training purposes, having no impact on conclusion.} and $b, c \sim \mathcal{U}(-1, 1)$ with fixed seed as input and calculated outputs accordingly. It's noted that the outputs might be complex numbers. The train/test/validation set is split as 0.8/0.1/0.1. We set the trained model as the reference model with $\epsilon=0.1$. 
All three types of models achieved promising results, shown in the caption of Fig. \ref{fig:illustration of model flexibility and explanation flexibility}, and first-order, and second-order explanation spaces are provided using swarm plots. 

\subsubsection{Ground truth feature attributions}
The given function $f^{*} = \frac{-b\pm\sqrt{b^2-4ac}}{2a}$ can be seen as an ideal model with $\E[L(f^{*}(\bfs{X}))] = 0$, which corresponds to a true feature interaction. The expected FIS of $a, b$ can be calculated as follows:
\begin{align*}
    FIS_{a,b}(f^{*}) = & \varphi_{a,b}(f^{*}) - (\varphi_{a}(f^{*}) + \varphi_{b}(f^{*})) \\ \nonumber
    = & \E[L(f^{*}(\bfs{X}_{\setminus \{a,b\}}))]  - \E[L(f^{*}(\bfs{X}_{\setminus \{a\}}))] -  \E[L(f^{*}(\bfs{X}_{\setminus \{b\}}))] \\ \nonumber
    \simeq & \sum_{i=1}^{n}(f^{*}(\bfs{X}_{\setminus \{a, b\}})- \bfs{y})^2 - \sum_{i=1}^{n}(f^{*}(\bfs{X}_{\setminus \{a\}})- \bfs{y})^2 - \sum_{i=1}^{n}(f^{*}(\bfs{X}_{\setminus \{b\}})- \bfs{y})^2,
\end{align*}
where $n$ is the number of samples and interaction between $b, c$ and $a, c$ can be calculated in the same way in the sampled dataset. The absolute interaction values indicate the strength, while the signs denote positive or negative interactions, as summarized in the Table. \ref{tab:gth}.  
\begin{table}[h!]
\caption{Ground truth feature attributions calculated from the given distribution. To reduce the influence of shuffling randomness, we calculated 100 times for each attribution and reported the average with standard error.}
\label{tab:gth}
\centering
\scriptsize
\begin{tabular}{ccccccc}
\toprule
Feature(s) & $\varphi_{a}(f^{*})$        & $\varphi_{b}(f^{*})$       & $\varphi_{c}(f^{*})$      & $|\varphi_{a, b}(f^{*})$      & $\varphi_{a, c}(f^{*})$     & $\varphi_{b, c}(f^{*})$     \\ \midrule
Attribution & $17.61 \pm 0.15$ & $13.52 \pm 0.20$ & $0.84 \pm 0.004$ & $-11.04 \pm 0.30$ & $-0.22 \pm 0.12$& $-0.18 \pm 0.30$ \\ \bottomrule
\end{tabular}
\end{table}

\subsubsection{The selection of the reference loss and model}
We can observe that RF is not trained as well as NNs and its reference attributions are not as accurate as the other two models. The selection of the loss function and the boundary definition within the Rashomon set are crucial. For example, one might quantify the Rashomon set using $R^2$ and set the boundary as R2 greater than 0.95, which would include all three models in the Rashomon set. Alternatively, if MAE is chosen as the metric with a maximum tolerance of 0.03, the RF would be excluded. The selection of the reference model depends on the specific user and task, or it may be provided as a black-box model.
Here we do not make any recommendations.
Not all current methods satisfy this condition, such as Trees FAst RashoMon Sets (TreeFARMS) that specifies trees as the model class~\citep{NEURIPS2022_5afaa8b4}.
We note that first-order, and second-order feature attributions calculated from our framework in the Rashomon set encompass the ground truth value, thereby decreasing the dependency of the reference model. This is because our generalized feature attribution score function is based on the reference loss. For example, although the loss of the RF model is greater than that of the other two models, their attribution values are similar.

\subsection{Statistical Comparison With Existing Methods}
On the existence of different Rashomon set sampling methods, feature attributions can be statistically quantified and compared, although ground truth normally does not exist in real-world problems. 
We mainly compared GRS with the following methods: variable importance cloud (VIC)~\citep{dong2020exploring}, which maps every variable to its importance for every good predictive model; TreeFARMS is the first technique for completely enumerating the Rashomon set for sparse decision trees; adversarial weight perturbations (AWP) and Random weights initialization~\citep{tsai2021formalizing} control weights for NNs sampling.

\subsubsection{Experimental settings}
We applied different sampling methods on 5 real-world datasets: COMPAS~\citep{dressel2018accuracy}, the Fair Isaac (FICO) credit risk dataset~\citep{FICO2018} used for the Explainable ML Challenge, and four coupon datasets (Bar, Coffee House, and Expensive Restaurant)~\citep{wang2017bayesian} with epsilons from [0.01, 0.03, 0.05, 0.1, 0.15]. To ensure a fair comparison, all sampled models are evaluated based on the same logistic loss and the same test set, by which all feature attributions are calculated accordingly. It is noted that while some sampled models might be included in their loss calculation, they are excluded from our evaluation. For each method, we followed their instructions to sample models with their code\citep{hsu2022rashomon, dong2020exploring, NEURIPS2022_5afaa8b4}. Different algorithms generate various reference models with different loss values, e.g., TreeFARMS utilizing trees; VIC utilizing logistic regressor; and others using MLPs. We select the minimum loss as the optimal loss among all trained models and benchmark other sampled Rashomon set. For random sampling and VIC, we sampled 200 and 1,000 models respectively, and selected valid models only, while treeFARMS automatically sampled sparse trees.  All results are included in the main text and Appendix.

\subsubsection{Feature attributions can be searched from different directions}

The FER, colored by SER, is calculated from the above sampled Rashomon set, where transparency indicates lower SER and solid color indicates higher validity, as shown in Fig. \ref{fig:fig5} (a), (b), and (c). The random searching generates some invalid models. 
Here is an example demonstrating the importance of evaluation metrics. A top-down searching algorithm, such as TreeFARMS, ensures search efficiency based on its pre-defined loss function. However, some models may fall outside the Rashomon set when benchmarked against the selected reference model, an MLP in this case, that achieves a lower loss value, demonstrating the influence of the reference model selection.

We can observe that TreeFARMS achieves a greater range of FER, as the aim of the algorithm is to find all sparse decision trees. This results in a substantial number of sampled models, thereby covering a wider range of FER. Our method finds the second-largest range in first-order feature attributions and the largest range in second-order feature attributions. 
Results from other datasets are included in the Appendix and they show that our framework can explore an acceptable range of FER compared with other methods. To ensure reliable model sampling and trustworthy explanations, it is crucial to select an appropriate reference model that incorporates generalizability and implementation sparsity. Without these, the risk of failing to sample models increases, as observed with VIC's inability to sample models using logistic classifiers for breast cancer and FICO datasets when epsilon is set as 0.01. Additionally, a poorly trained reference model,  e.g., TreeFARMS on the breast cancer dataset, can lead to untrustworthy explanations.

We also illustrate the variance of different methods on a single epsilon in Fig. \ref{fig:fig5} (d) and (e), and the variance of the same method across different epsilons in Fig. \ref{fig:COMPAS-MR-ALL-E} and \ref{fig:COMPAS-FIS-ALL-E} separately. The trends in attributions show coherence across different methods, as demonstrated by the results from the COMPAS dataset. These results consistently indicate that the absence of juvenile felonies and misdemeanors correlates with a lower likelihood of recidivism within two years. Conversely, a history of previous criminal activity, especially with more than three offenses, is associated with a higher likelihood of reoffending. From the second-order attributions, we focus on interactions involving teenagers (age less than 26). Interestingly, teenagers younger than 23 with no juvenile crimes, or with 2-3 felony offenses, have a higher risk of committing crimes. In contrast, those who have committed only one crime are less likely to make mistakes again. According to results from VIC \citep{dong2020exploring}, gender becomes more important than age. However, both AWP and random sampling methods explore only parts of the FER for Sex and Age, and their overall importance can be similar, as some methods might be stuck at local FER. Therefore, it is crucial to consider the overall FER across various methods, including all valid models, requiring further work.

\subsubsection{Most models are redundant in terms of attributions}
In Fig. \ref{fig:fig5} (a), the expected number of models to explore varies across datasets and methodologies, and some generated models are not within the Rashomon set defined by the practical optimal loss. We included the relevant epsilon values in legends of detailed feature attribution plots, e.g., in Fig. \ref{fig:COMPAS-MR-ALL-E} and \ref{fig:COMPAS-FIS-ALL-E}. This is why we discussed the generalized Rashomon set and did not force a model as optimally as possible. Methods such as random sampling, VIC, and TreeFARMS do not allow users to control the number of models. The random sampling method generates most models outside the Rashomon set, but with higher epsilons, the chance of valid models increases. Notably, we did not include all models in the plot due to computational constraints. For instance, in the case of TreeFARMS applied to the COMPAS dataset\footnote{COMPAS: We use the same discretized binary features of COMPAS produced in \citep{hu2019optimal}, which are the following: $\text{sex} = \text{Female}, \text{age} < 21, \text{age} < 23, \text{age} < 26, \text{age} < 46, \text{juvenile felonies} = 0, \text{juvenile misdemeanors} = 0, \text{juvenile crimes} = 0, \text{priors} = 0, \text{priors} = 1, \text{priors} = 2 \text{ to } 3, \text{priors} > 3$.}, it generates over 20,000 trees when the epsilon is 0.1, rendering it computationally impractical to consider all trees. 
Functionally, although the random sampling method generates less valid models than AWP, they explored similar FER. From Fig. \ref{fig:fig5} (b) and \ref{fig:fig5} (c), we can observe that VIC's FER in the first-order and the second-order attribution is generally consistent, not increasing with greater epsilons. Furthermore, VIC sampled more models than GRS, yet its FER does not exhibit a consistent improvement. Generally, FER increases with greater epsilons, as the greater tolerance allows features to contribute more to the predictions. However, FER does not increase with the number of sampled models. Results from other datasets are provided in the Appendix.
Results from other datasets are shown in the Appendix.

\begin{figure}[th!]
     \centering
     \begin{subfigure}{\textwidth}
         \centering
         \includegraphics[width=\textwidth]{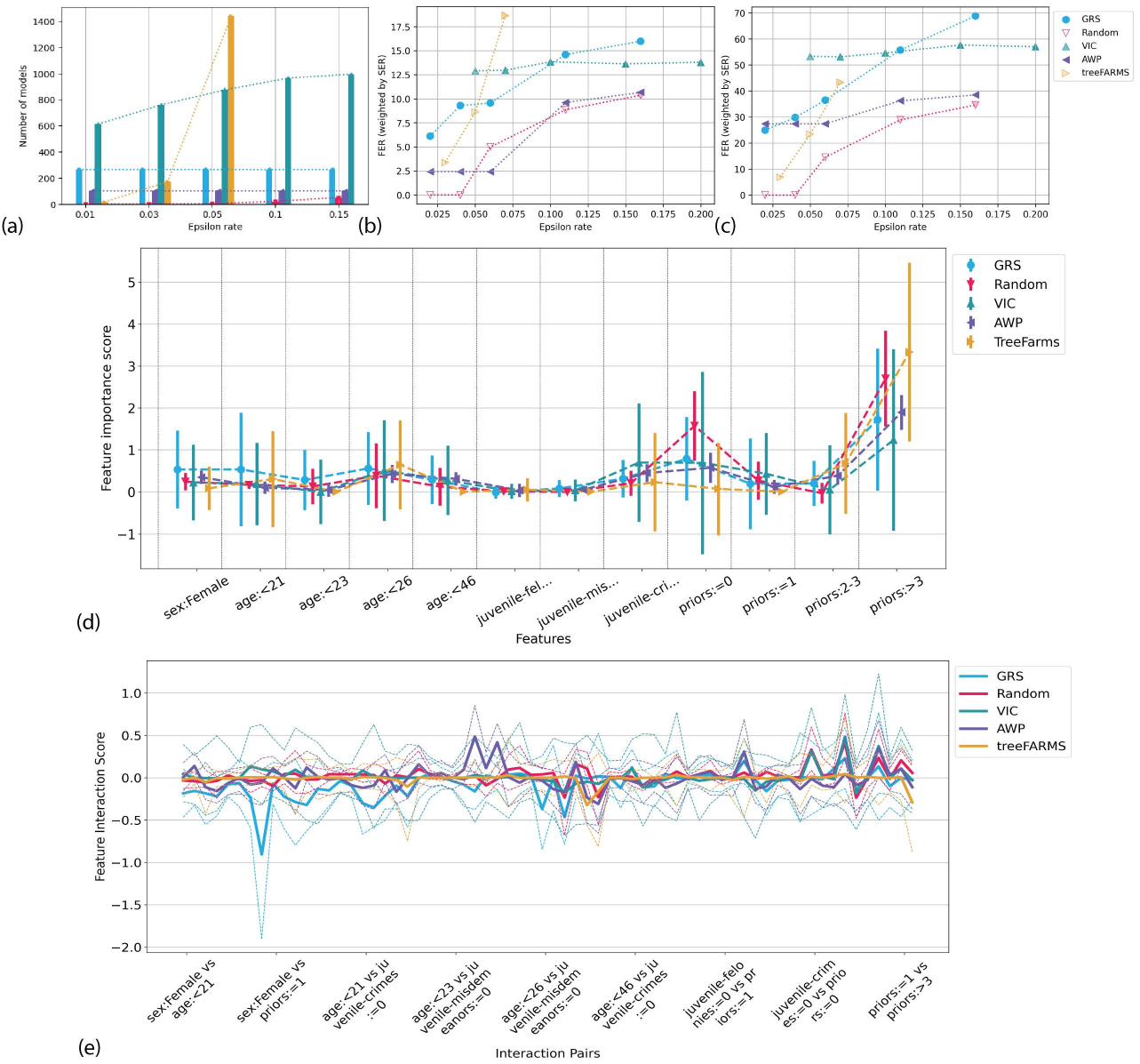}
     \end{subfigure}   
    \caption{Summary of the Rashomon subset from a set of epsilons on the dataset COMPAS, including (a) the number of sampled models on different methods, where no bar indicates too many models (greater than 20,000 models) (b) the first-order FER, where the $x$-axis is epsilon (benchmarking against optimal loss) and colors represent according to SER, (c) the second-order FER, following the same format as above, (d) the detailed first-order feature attribution on individual features when epsilon is set as 0.05, where the vertical bars represent the bounds and the dotted lines connect the average scores, and (e) the detailed second-order feature attribution on feature pairs when epsilon is set 0.05, where the dotted lines represent the bounds and the solid lines connect the average scores.
    Each color corresponds to a sampling method and due to space limitations, some interaction pairs are omitted on the $x$-axis.} 
\label{fig:fig5}
\end{figure}

\begin{figure}[h!]
     \centering
     \begin{subfigure}{\textwidth}
         \centering
         \includegraphics[width=\textwidth]{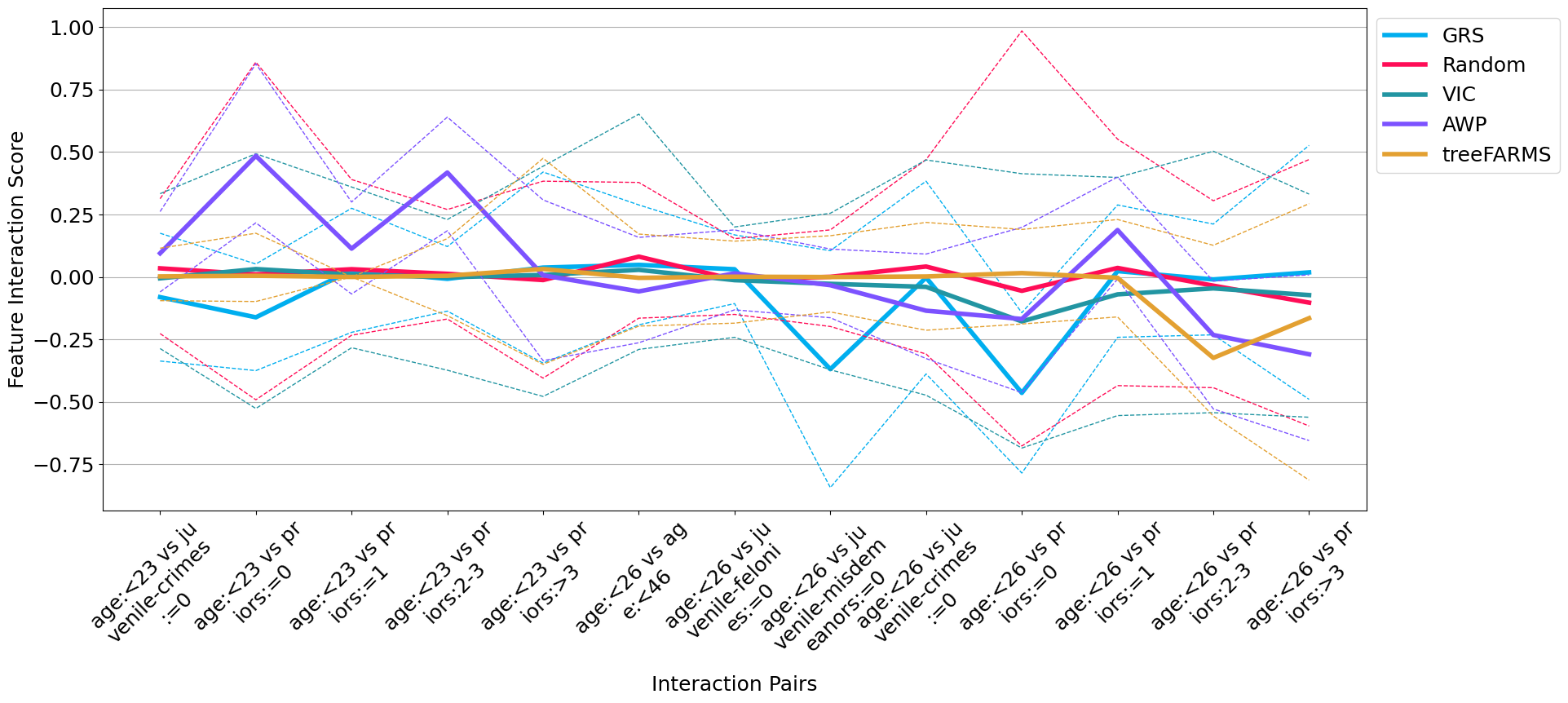}
     \end{subfigure}   
    \caption{The detailed second-order feature attribution on interested feature pairs.} 
\label{fig:fig5}
\end{figure}

\begin{figure}[h!]
     \centering
     \begin{subfigure}{\textwidth}
         \centering
         \includegraphics[width=.8\textwidth]{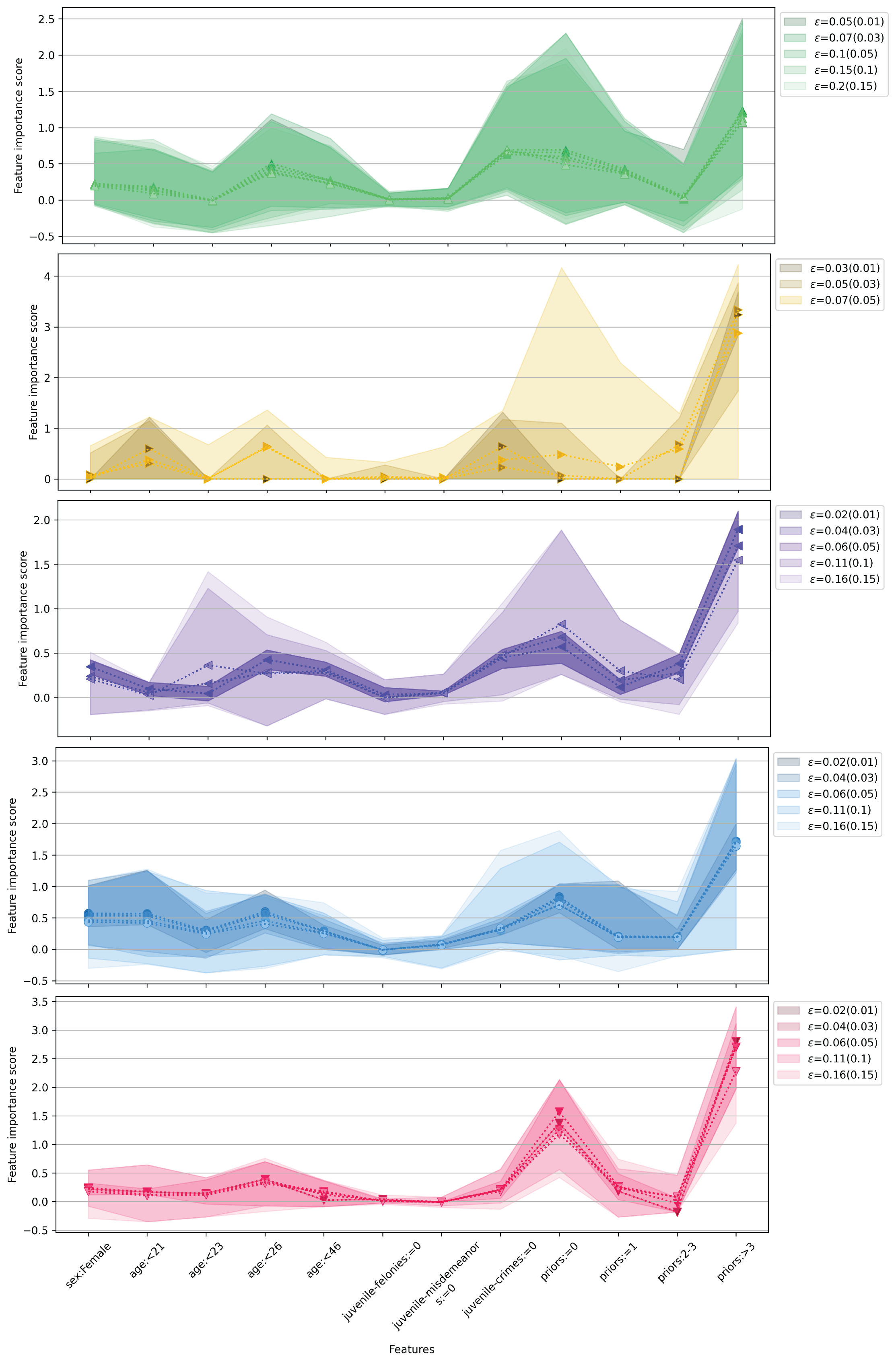}
     \end{subfigure}   
    \caption{Summary of detailed first-order feature attributions across different methods and epsilons on the dataset COMPAS, where the $x$-axis represents features and the $y$-axis displays feature importance scores. The legend includes epsilons relative to the reference model (in brackets) and to the optimal model identified in practice.} 
\label{fig:COMPAS-MR-ALL-E}
\end{figure}

\begin{figure}[h]
     \centering
     \begin{subfigure}{\textwidth}
         \centering
         \includegraphics[width=.8\textwidth]{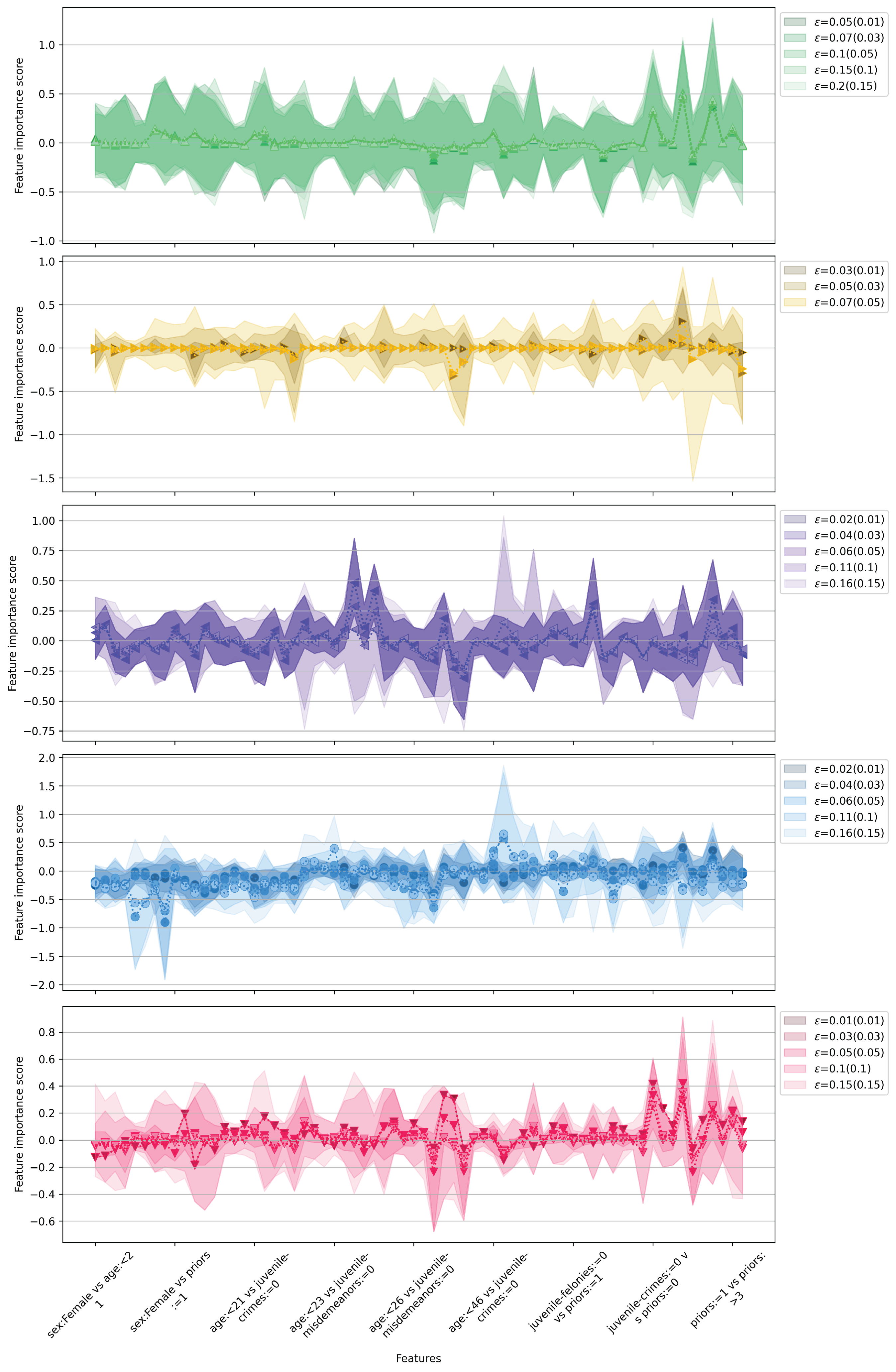}
     \end{subfigure}   
    \caption{Summary of detailed second-order feature attributions across different methods and epsilons on the dataset COMPAS, where the $x$-axis represents features and the $y$-axis displays feature importance scores. The legend includes epsilons relative to the reference model (in brackets) and to the optimal model identified in practice.} 
\label{fig:COMPAS-FIS-ALL-E}
\end{figure}

\begin{table}[]
\caption{\label{tab:table2}Summary of comparison for different Rashomon set exploring methods}
\centering
\scriptsize
\begin{tabular}{lp{3cm}p{2cm}p{1.8cm}p{1.8cm}}
\toprule
\textbf{Methods}   & \textbf{Model structure generalizability} & \textbf{Task type} & \textbf{Searching efficiency} & \textbf{Explanation flexibility}  \\ \midrule
\textbf{TreeFARMS} & Tree-based  & Classification                & True                    & First-order                                \\
\textbf{VIC}                & Logistic Classifier               & Classification                    & False                         & First-order                     \\
\textbf{GRS}                & ANY                & Classification \& Regression                    & True                          & High-order                     \\
\textbf{Random}             & NN-based                & Classification \& Regression                    & False                          & First-order                     \\
\textbf{AWP}                & NN-based                & Classification \& Regression                    & True                          & First-order                       \\ \bottomrule
\end{tabular}
\end{table}

\section{Conclusion}\label{sec:conclusion}
In this paper, our primary focus has been on the practical application of Rashomon sets, recognizing the current lack of guidance that hinders its broader adoption and impact across different fields. We identified two fundamental principles, generalizability and implementation sparsity, that sampling methods should follow (or at least consider). When comparing various sampling methods, it is essential that sampled Rashomon sets meet the criteria of model evaluation and feature attribution generalizability, although model structure generalizability may vary. The choice of the reference model can significantly influence these considerations and should be decided on a case-by-case basis. For practical utilization of a Rashomon set, we introduced searching efficiency and functional sparsity, aiming to identify non-redundant and effective models.
To address these requirements, we proposed an $\epsilon$-subgradient-based sampling framework that incorporates generalized feature attribution and statistical quantification. We validated this framework using both synthetic and real-world datasets, summarizing its desirable properties in  Table \ref{tab:table2}.
The potential impact of the Rashomon set and high-order feature attributions is extensive. Explanations play a crucial role in meeting transparency requirements across various domains, facilitating collaboration between humans and AI, and supporting model development, debugging, and monitoring efforts. XAI is increasingly integrated into scientific domains such as chemistry, biology, and physics, where thorough explanations are essential for reliable and informed decision-making.

\clearpage

\section*{Appendix}
\subsection*{Convergence proof}
There are many results on the convergence of the subgradient method. For constant step size and constant step length, the subgradient algorithm is guaranteed to converge to within some range of the optimal value.
For the diminishing step size rule (and therefore also the square summable but not
summable step size rule), the algorithm is guaranteed to converge to the optimal value.

We assume that there is a maximizer of the function and $\Loss(f_{ref}(\tilde{\bfs{X}})) = \Loss(f_{ref}(\bfs{X}^{*})) + \epsilon$. There are many results on the convergence of the subgradient method. For constant step size and constant step length, the subgradient algorithm is guaranteed to converge within
some range of the optimal value, and we have
\begin{equation}
    \lim_{k \to \infty} \Loss(f_{ref}(\bfs{X}^{(k)})) = \Loss(f_{ref}(\tilde{\bfs{X}}))   
\end{equation}

\begin{align}
    \| \bfs{X}^{*} - \bfs{X}^{(k+1)} \|_{2}^{2} &=  \| \bfs{X}^{*} - \bfs{X^{(k)}} - \lambda_{k} \bfs{z}^{(k)} \|_{2}^{2} \\
    & = \| \bfs{X}^{*} - \bfs{X}^{(k)} \|_{2}^{2} - 2 \lambda_{k}\bfs{z}^{(k)T}(\bfs{X}^{*} - \bfs{X}^{(k)}) + \lambda_{k}^{2}\|\bfs{z}^{(k)} \|_{2}^{2} \\
    & \geq \| \bfs{X}^{*} - \bfs{X}^{(k)} \|_{2}^{2} - 2 \lambda_{k}\bfs{z}^{(k)T}(\Loss_{ref}(\bfs{X}^{*}) - \Loss_{ref}(\bfs{X}^{(k)})) + \lambda_{k}^{2}\|\bfs{z}^{(k)} \|_{2}^{2} \\
    & \geq \| \bfs{X}^{*} - \bfs{X}^{(k)} \|_{2}^{2} - 2 \lambda_{k}\bfs{z}^{(k)T}(\Loss_{ref}(\bfs{X}^{*}) - \Loss_{ref}(\bfs{X}^{(k)}) + \epsilon) + \lambda_{k}^{2}\|\bfs{z}^{(k)} \|_{2}^{2}.
\end{align}

\noindent Applying the inequality above recursively, we have
\begin{equation}
    \| \bfs{X}^{*} - \bfs{X}^{(k+1)} \|_{2}^{2} \geq 
    \| \bfs{X}^{*} - \bfs{X}^{(1)} \|_{2}^{2} - 2 \sum_{i=1}^{k}\lambda_{i}(\Loss_{ref}(\bfs{X}^{*}) - \Loss_{ref}(\bfs{X}^{(i)}) + \epsilon) + \sum_{i=1}^{k}\lambda_{i}^{2}\|\bfs{z}^{(i)} \|_{2}^{2}.
\end{equation}

\noindent Using $\| \bfs{X}^{*} - \bfs{X}^{(k+1)} \|_{2}^{2} \geq 0$ we have 
\begin{equation}
    2 \sum_{i=1}^{k} \lambda_{i}(\Loss_{ref}(\bfs{X}^{*}) - \Loss_{ref}(\bfs{X}^{i}) + \epsilon) \geq \| \bfs{X}^{*} - \bfs{X}^{(1)} \|_{2}^{2} + \sum_{i=1}^{k}\lambda_{i}^{2}\|\bfs{z}^{(i)} \|_{2}^{2}.
\end{equation}

\noindent Combining with $ \sup_{i=1,2,...,k} \Loss_{ref}(\bfs{X}^{(i)}) = \Loss_{ref}(\bfs{X}^{*}) + \epsilon$ we have the inequality
\begin{equation}
    \sum_{i=1}^{k} \lambda_{i}(\sup_{i=1,2,..,k}(\Loss_{ref}(\bfs{X}^{(i)})) - \Loss_{ref}(\bfs{X}^{(i)}))
    \geq 
    \frac{\| \bfs{X}^{*} - \bfs{X}^{(1)} \|_{2}^{2} + \sum_{i=1}^{k}\lambda_{i}^{2}\|\bfs{z}^{(i)} \|_{2}^{2}}{2 \sum_{i=1}^{k} \lambda_{i}}.
\end{equation}
\noindent This can be reformulated as
\begin{equation}
    \sum_{i=1}^{k} \lambda_{i}(\Loss_{ref}(\bfs{X}^{(i)}) - \sup_{i=1,2,..,k}(\Loss_{ref}(\bfs{X}^{(i)})))
    \leq 
    \frac{ - \| \bfs{X}^{*} - \bfs{X}^{(1)} \|_{2}^{2} - \sum_{i=1}^{k}\lambda_{i}^{2}\|\bfs{z}^{(i)} \|_{2}^{2}}{2 \sum_{i=1}^{k} \lambda_{i}}.
\end{equation}

\noindent Here we introduce the Rashomon condition and start searching from $\bfs{X}^{(1)} = \bfs{X}^{*}$, then we have 
\begin{equation}
    \sum_{i=1}^{k} (\Loss_{ref}(\bfs{X}^{(i)}) - \sup_{i=1,2,..,k}(\Loss_{ref}(\bfs{X}^{(i)})))  \leq -\frac{1}{2}\sum_{i=1}^{k}\lambda_{i}\|\bfs{z}^{(i)} \|_{2}^{2}.
\end{equation}

As $k \to \infty$, the right side of the inequality $\sum_{i=1}^{k}\lambda_{i}\|\bfs{z}^{(i)} \|_{2}^{2}$ converges to zero. In other words, the subgradient method converges $$\lim_{k \to \infty} \Loss(f_{ref}(\bfs{X}^{(k)})) \to \sup_{i=1,2,..,k}(\Loss_{ref}(\bfs{X}^{(i)})) \to \Loss(f_{ref}(\bfs{X}^{*})) + \epsilon.$$

\subsection*{Additional Results}

The selected binary features for other datasets \citep{FICO2018, hu2019optimal, wang2017bayesian, NEURIPS2022_5afaa8b4} are listed below. 

\begin{itemize}
    \item \textbf{Bar}: The selected binary features are ``Bar = 1 to 3'', ``Bar = 4 to 8'', ``Bar = less1'', ``maritalStatus = Single'', ``childrenNumber = 0'', ``Bar = gt8'', ``passanger = Friend(s)'', ``time = 6PM'', ``passanger = Kid(s)'', ``CarryAway = 4 to 8'', ``gender = Female'', ``education = Graduate degree (Masters Doctorate etc.)'', ``Restaurant20To50 = 4 to 8'', ``expiration = 1d'', ``temperature = 55''.
    
    \item \textbf{Coffee House}: The selected binary features are ``CoffeeHouse = 1 to 3'', ``CoffeeHouse = 4 to 8'', ``CoffeeHouse = gt8'', ``CoffeeHouse = less1'', ``expiration = 1d'', ``destination = No Urgent Place'', ``time = 10AM'', ``direction = same'', ``destination = Home'', ``toCoupon = GEQ15min'', ``Restaurant20To50 = gt8'', ``education = Bachelors degree'', ``time = 10PM'', ``income = \$75000 - \$87499'', ``passanger = Friend(s)''.
    
    \item \textbf{Expensive Restaurant}: The selected binary features are ``expiration = 1d'', ``CoffeeHouse = 1 to 3'', ``Restaurant20To50 = 4 to 8'', ``Restaurant20To50 = 1 to 3'', ``occupation = Office \& Administrative Support'', ``age = 31'', ``Restaurant20To50 = gt8'', ``income = \$12500 - \$24999'', ``toCoupon = GEQ15min'', ``occupation = Computer \& Mathematical'', ``time = 10PM'', ``CoffeeHouse = 4 to 8'', ``income = \$50000 - \$62499'', ``passanger = Alone'', ``destination = No Urgent Place''.
    
    \item \textbf{Breast Cancer} The selected binary features are ``Clump\_Thickness = 10'', ``Uniformity\_Cell\_Size = 1'', ``Uniformity\_Cell\_Size = 10'', ``Uniformity\_Cell\_Shape = 1'', ``Marginal\_Adhesion = 1'', ``Single\_Epithelial\_Cell\_Size = 2'', ``Bare\_Nuclei = 1'', ``Bare\_Nuclei = 10'', ``Normal\_Nucleoli = 1'', ``Normal\_Nucleoli = 10''.

    \item \textbf{FICO}: The selected binary features are ``External Risk Estimate $<$ 0.49'', ``External Risk Estimate $<$ 0.65'', ``External Risk Estimate $<$ 0.80'', ``Number of Satisfactory Trades $<$ 0.5'', ``Trade Open Time $<$ 0.6'', ``Trade Open Time $<$ 0.85'', ``Trade Frequency $<$ 0.45'', ``Trade Frequency $<$ 0.6'', ``Delinquency $<$ 0.55'', ``Delinquency $<$ 0.75'', ``Installment $<$ 0.5'', ``Installment $<$ 0.7'', ``Inquiry $<$ 0.75'', ``Revolving Balance $<$ 0.4'', ``Revolving Balance $<$ 0.6'', ``Utilization $<$ 0.6'', ``Trade W. Balance $<$ 0.33''.
\end{itemize}

\begin{figure}[h!]
     \centering
     \begin{subfigure}{\textwidth}
         \centering
         \includegraphics[width=\textwidth]{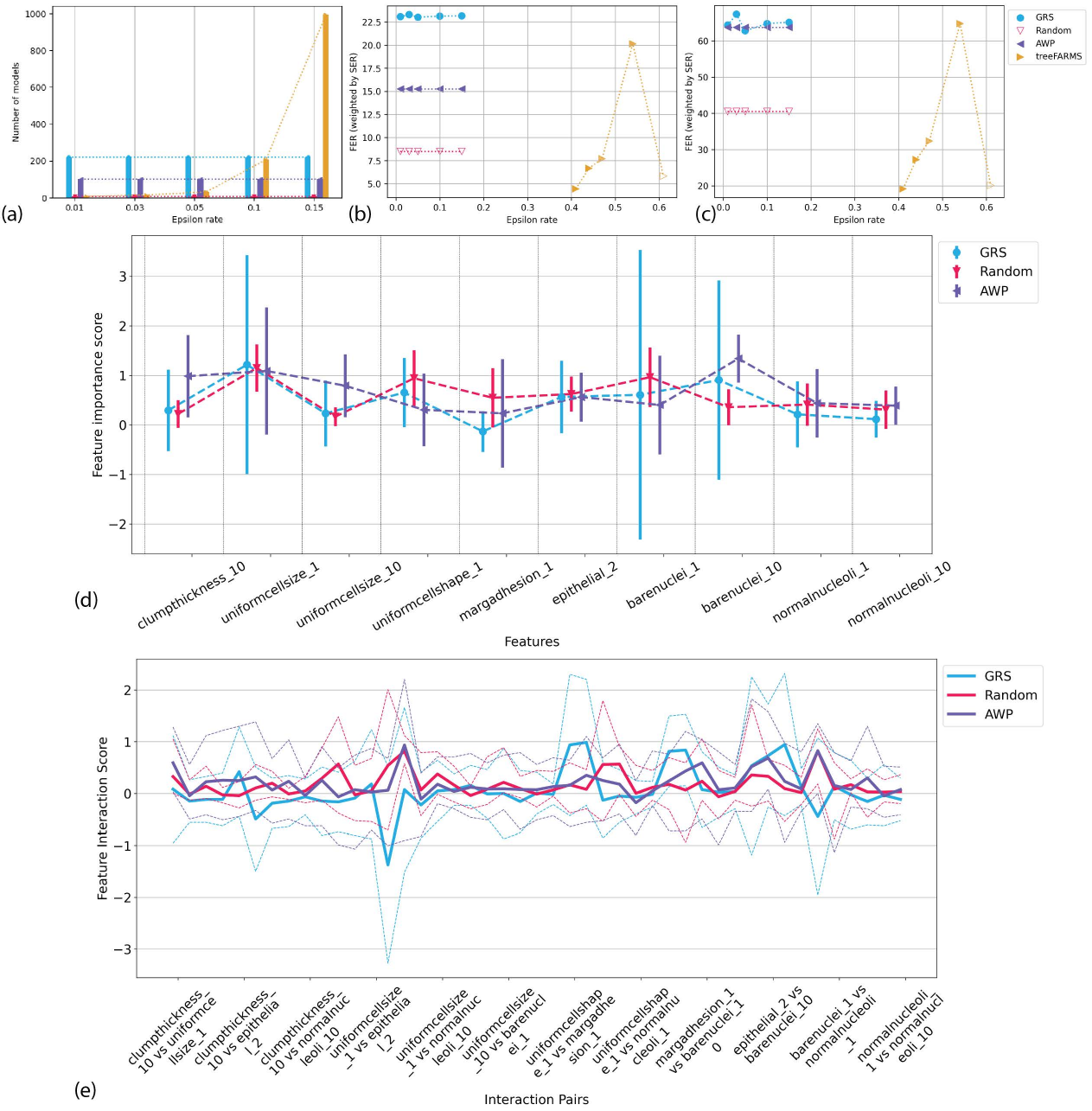}
         \label{fig:fig3}
     \end{subfigure}   
    \caption{Summary of the Rashomon subset from a set of epsilons on the dataset breast cancer, including (a) the number of sampled models on different methods, where no bar indicates too many models (greater than 20,000 models), (b) the first-order FER, where the $x$-axis is epsilon (benchmarking against optimal loss) and colors represent according to SER, (c) the second-order FER, following the same format as above, (d) the detailed first-order feature attribution on individual features when epsilon is set as 0.05, where the vertical bars represent the bounds and the dotted lines connect the average scores, and (e) the detailed second-order feature attribution on feature pairs when epsilon is set 0.05, where the dotted lines represent the bounds and the solid lines connect the average scores.
    Each color corresponds to a sampling method and due to space limitations, some interaction pairs are omitted on the $x$-axis.} 
\label{fig:explanation space summary}
\end{figure}
\begin{figure}[h!]
     \centering
     \begin{subfigure}{\textwidth}
         \centering
         \includegraphics[width=\textwidth]{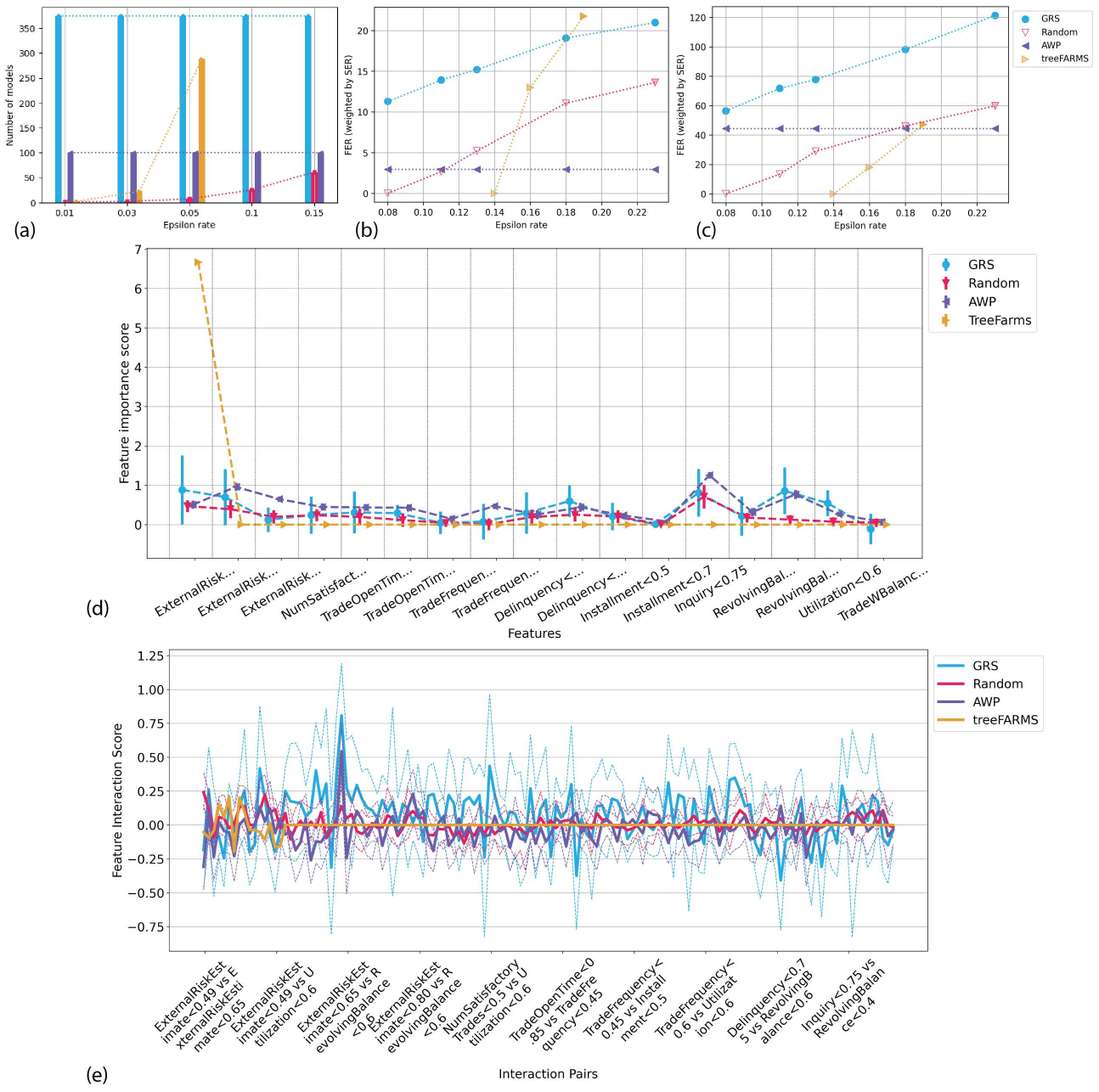}
         \label{fig:fig3}
     \end{subfigure}   
    \caption{Summary of the Rashomon subset from a set of epsilons on the dataset FICO, including (a) the number of sampled models on different methods, where no bar indicates too many models (greater than 20,000 models), (b) the first-order FER, where the $x$-axis is epsilon (benchmarking against optimal loss) and colors represent according to SER, (c) the second-order FER, following the same format as above, (d) the detailed first-order feature attribution on individual features when epsilon is set as 0.05, where the vertical bars represent the bounds and the dotted lines connect the average scores, and (e) the detailed second-order feature attribution on feature pairs when epsilon is set 0.05, where the dotted lines represent the bounds and the solid lines connect the average scores.
    Each color corresponds to a sampling method and due to space limitations, some interaction pairs are omitted on the $x$-axis.} 
\label{fig:explanation space summary fico}
\end{figure}
\begin{figure}[h!]
     \centering
     \begin{subfigure}{\textwidth}
         \centering
         \includegraphics[width=\textwidth]{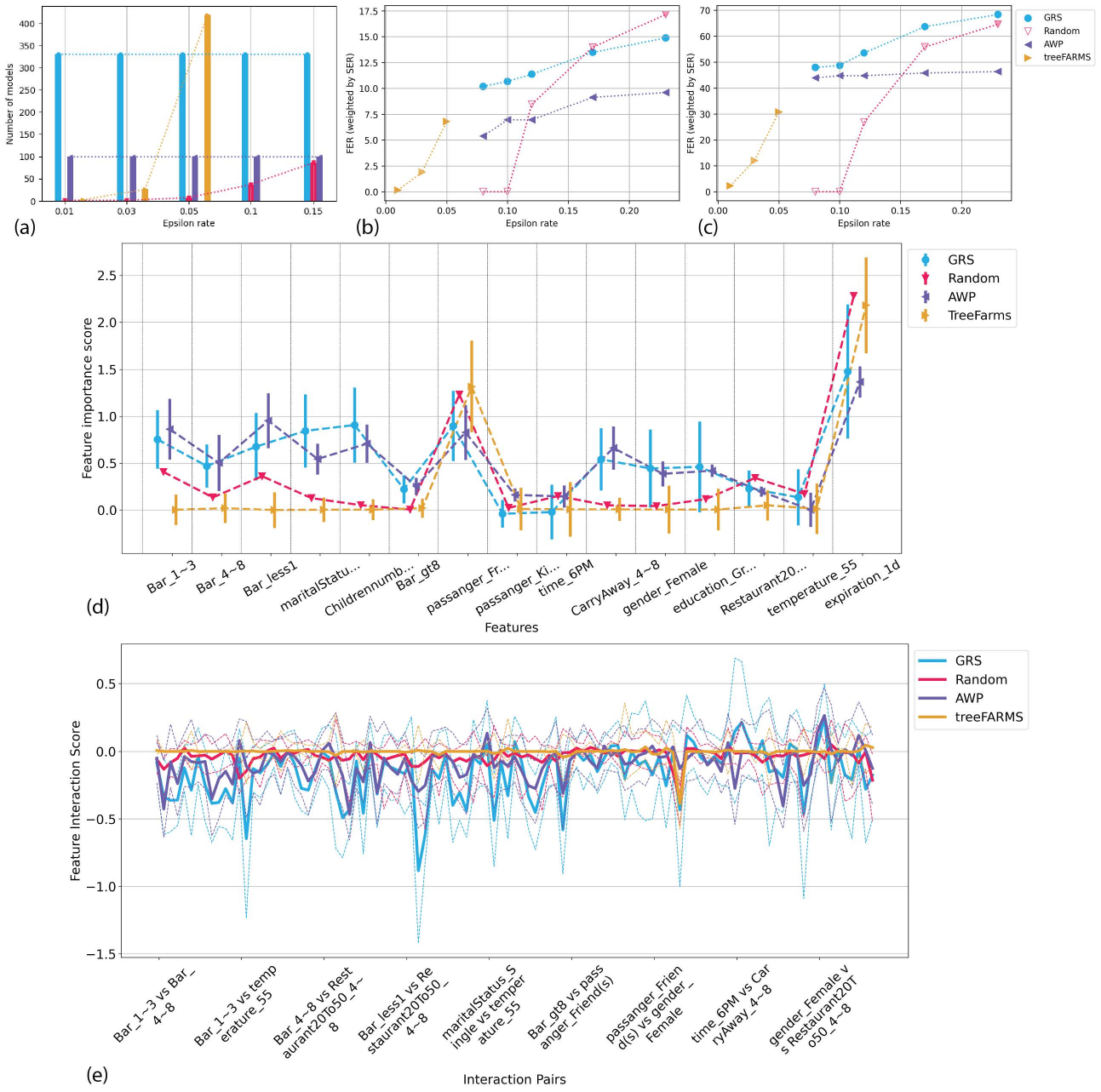}
     \end{subfigure}   
    \caption{Summary of the Rashomon subset from a set of epsilons on the dataset BAR, including (a) the number of sampled models on different methods, where no bar indicates too many models (greater than 20,000 models) (b) the first-order FER, where the $x$-axis is epsilon (benchmarking against optimal loss) and colors represent according to SER, (c) the second-order FER, following the same format as above, (d) the detailed first-order feature attribution on individual features when epsilon is set as 0.05, where the vertical bars represent the bounds and the dotted lines connect the average scores, and (e) the detailed second-order feature attribution on feature pairs when epsilon is set 0.05, where the dotted lines represent the bounds and the solid lines connect the average scores.
    Each color corresponds to a sampling method and due to space limitations, some interaction pairs are omitted on the $x$-axis.} 
\label{fig:explanation space summary bar}
\end{figure}
\begin{figure}[h!]
     \centering
     \begin{subfigure}{\textwidth}
         \centering
         \includegraphics[width=\textwidth]{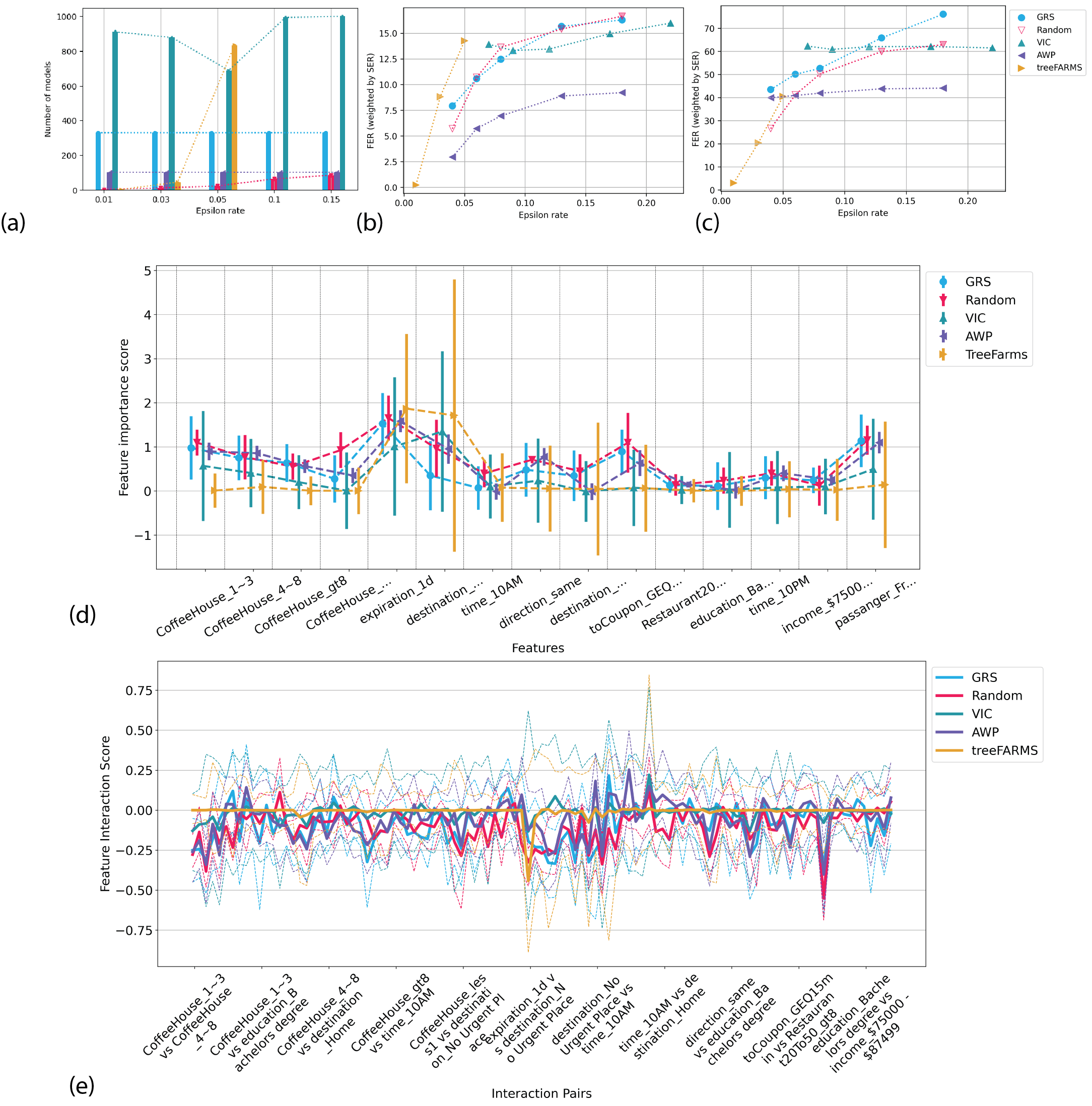}
     \end{subfigure}   
    \caption{Summary of the Rashomon subset from a set of epsilons on the dataset coffee house, including (a) the number of sampled models on different methods, where no bar indicates too many models (greater than 20,000 models) (b) the first-order FER, where the $x$-axis is epsilon (benchmarking against optimal loss) and colors represent according to SER, (c) the second-order FER, following the same format as above, (d) the detailed first-order feature attribution on individual features when epsilon is set as 0.05, where the vertical bars represent the bounds and the dotted lines connect the average scores, and (e) the detailed second-order feature attribution on feature pairs when epsilon is set 0.05, where the dotted lines represent the bounds and the solid lines connect the average scores.
    Each color corresponds to a sampling method and due to space limitations, some interaction pairs are omitted on the $x$-axis.} 
\label{fig:explanation space summary coffee house}
\end{figure}
\begin{figure}[h!]
     \centering
     \begin{subfigure}{\textwidth}
         \centering
         \includegraphics[width=\textwidth]{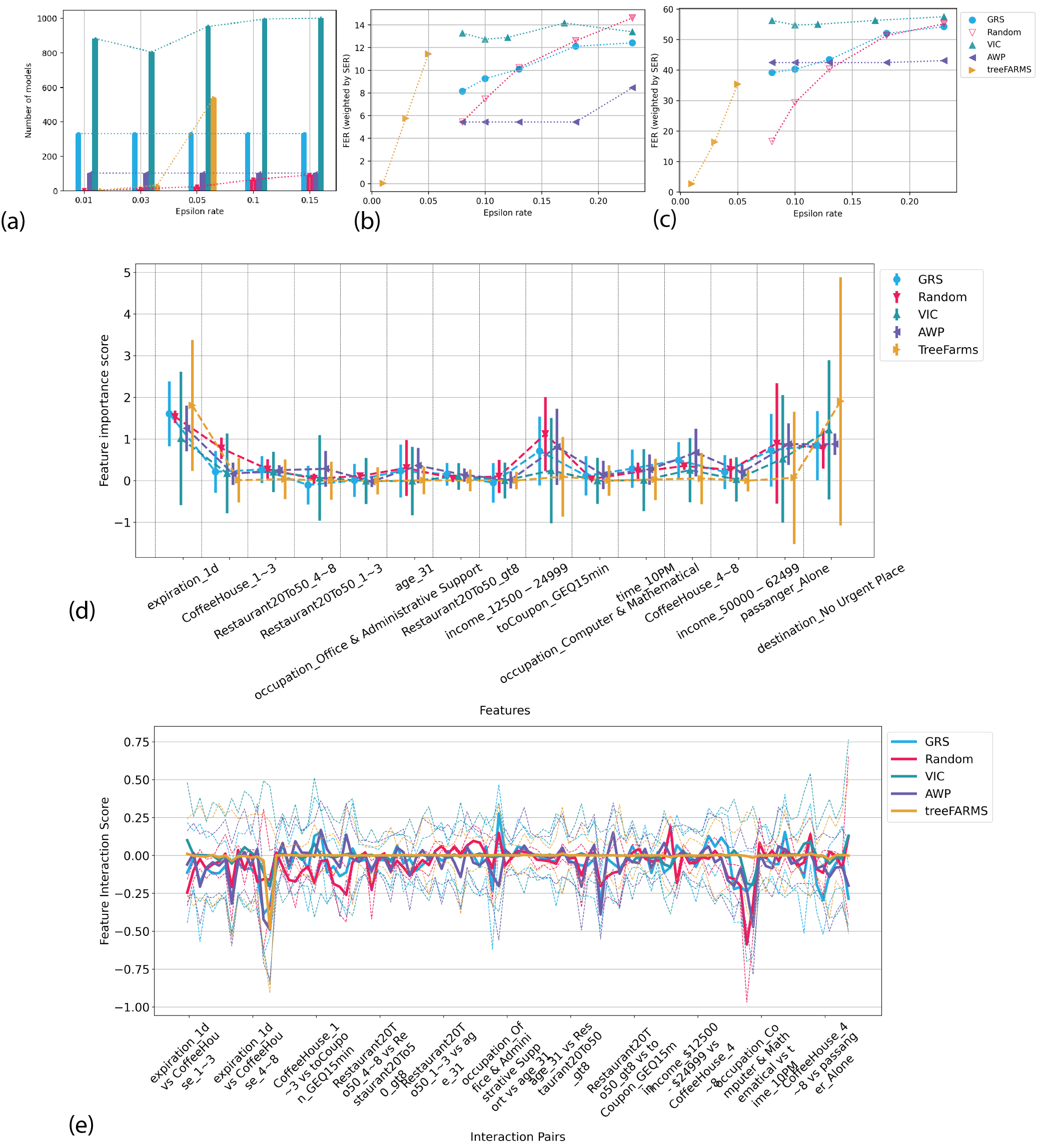}
     \end{subfigure}   
    \caption{Summary of the Rashomon subset from a set of epsilons on the dataset expensive restaurant, including (a) the number of sampled models on different methods, where no bar indicates too many models (greater than 20,000 models) (b) the first-order FER, where the $x$-axis is epsilon (benchmarking against optimal loss) and colors represent according to SER, (c) the second-order FER, following the same format as above, (d) the detailed first-order feature attribution on individual features when epsilon is set as 0.05, where the vertical bars represent the bounds and the dotted lines connect the average scores, and (e) the detailed second-order feature attribution on feature pairs when epsilon is set 0.05, where the dotted lines represent the bounds and the solid lines connect the average scores.
    Each color corresponds to a sampling method and due to space limitations, some interaction pairs are omitted on the $x$-axis.} 
\label{fig:explanation space summary expensive restaurant}
\end{figure}

\begin{figure}[h!]
     \centering
     \begin{subfigure}{\textwidth}
         \centering
         \includegraphics[width=\textwidth]{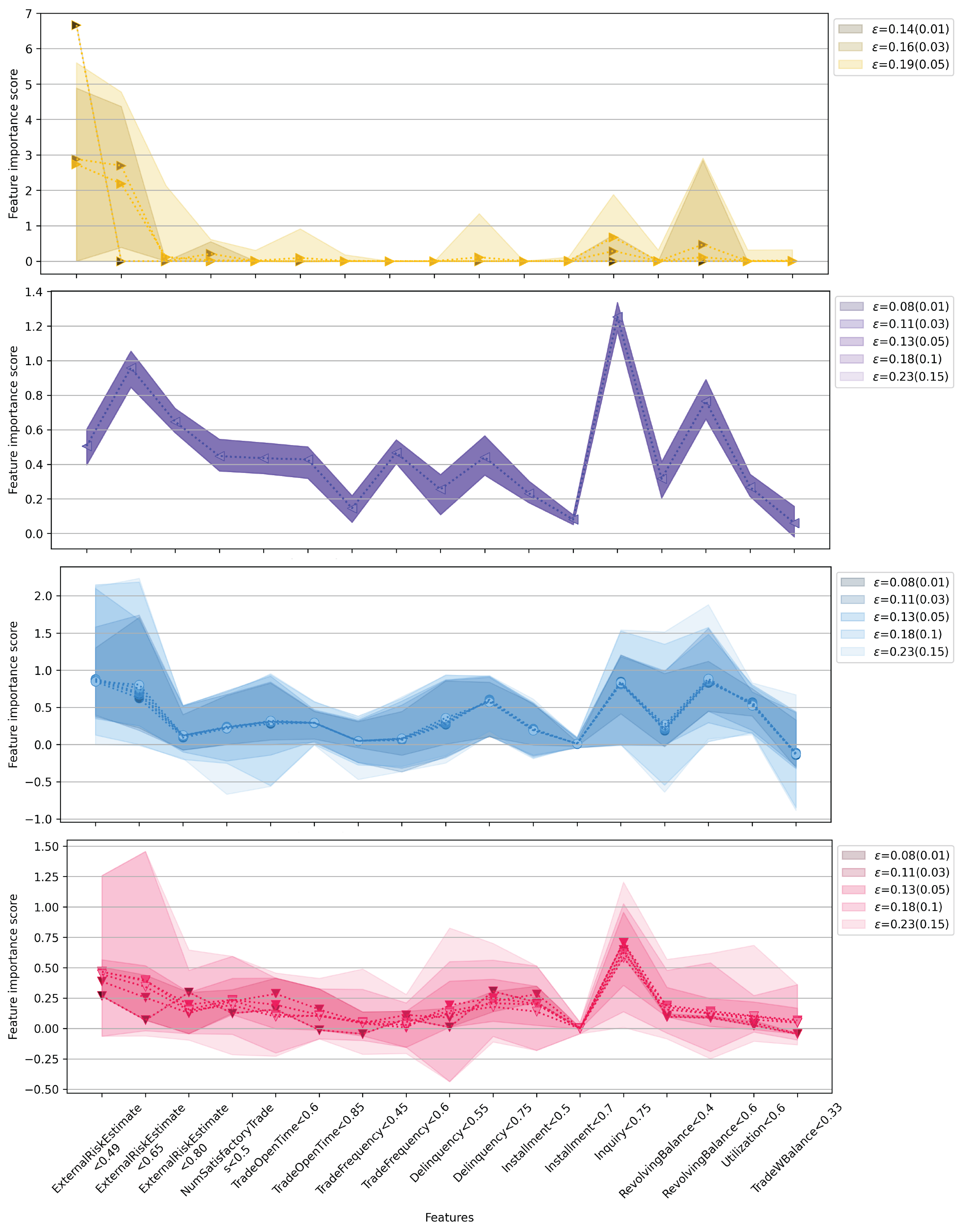}
     \end{subfigure}   
    \caption{Summary of detailed first-order feature attributions across different methods and epsilons on the dataset FICO, where the $x$-axis represents features and the $y$-axis displays feature importance scores. The legend includes epsilons relative to the reference model (in brackets) and to the optimal model identified in practice.} 
\label{fig:FICO-MR-ALL-E}
\end{figure}
\begin{figure}[h!]
     \centering
     \begin{subfigure}{\textwidth}
         \centering
         \includegraphics[width=\textwidth]{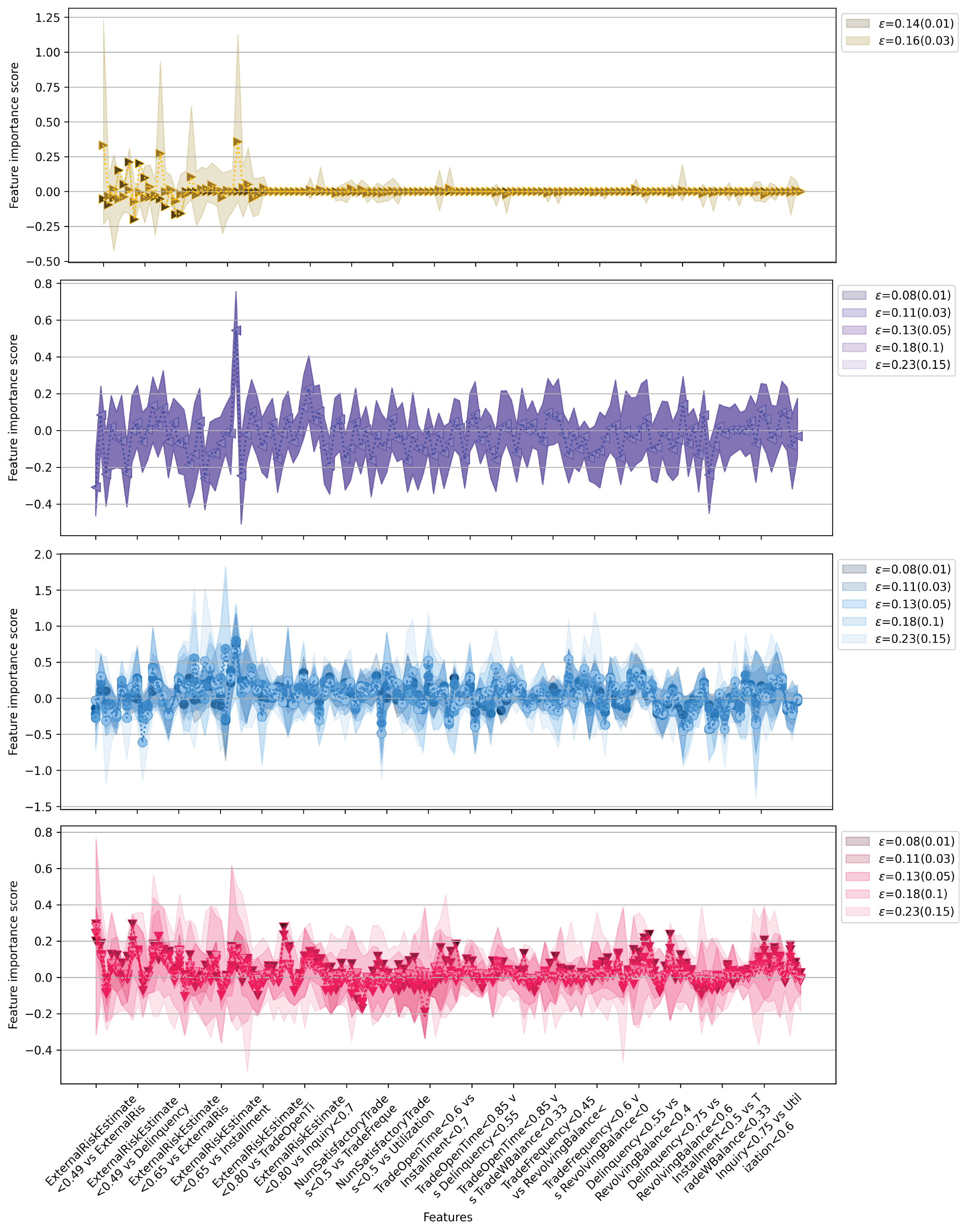}
     \end{subfigure}   
    \caption{Summary of detailed second-order feature attributions across different methods and epsilons on the dataset FICO, where the $x$-axis represents features and the $y$-axis displays feature importance scores. The legend includes epsilons relative to the reference model (in brackets) and to the optimal model identified in practice.} 
\label{fig:FICO-FIS-ALL-E}
\end{figure}

\begin{figure}[h!]
     \centering
     \begin{subfigure}{\textwidth}
         \centering
         \includegraphics[width=\textwidth]{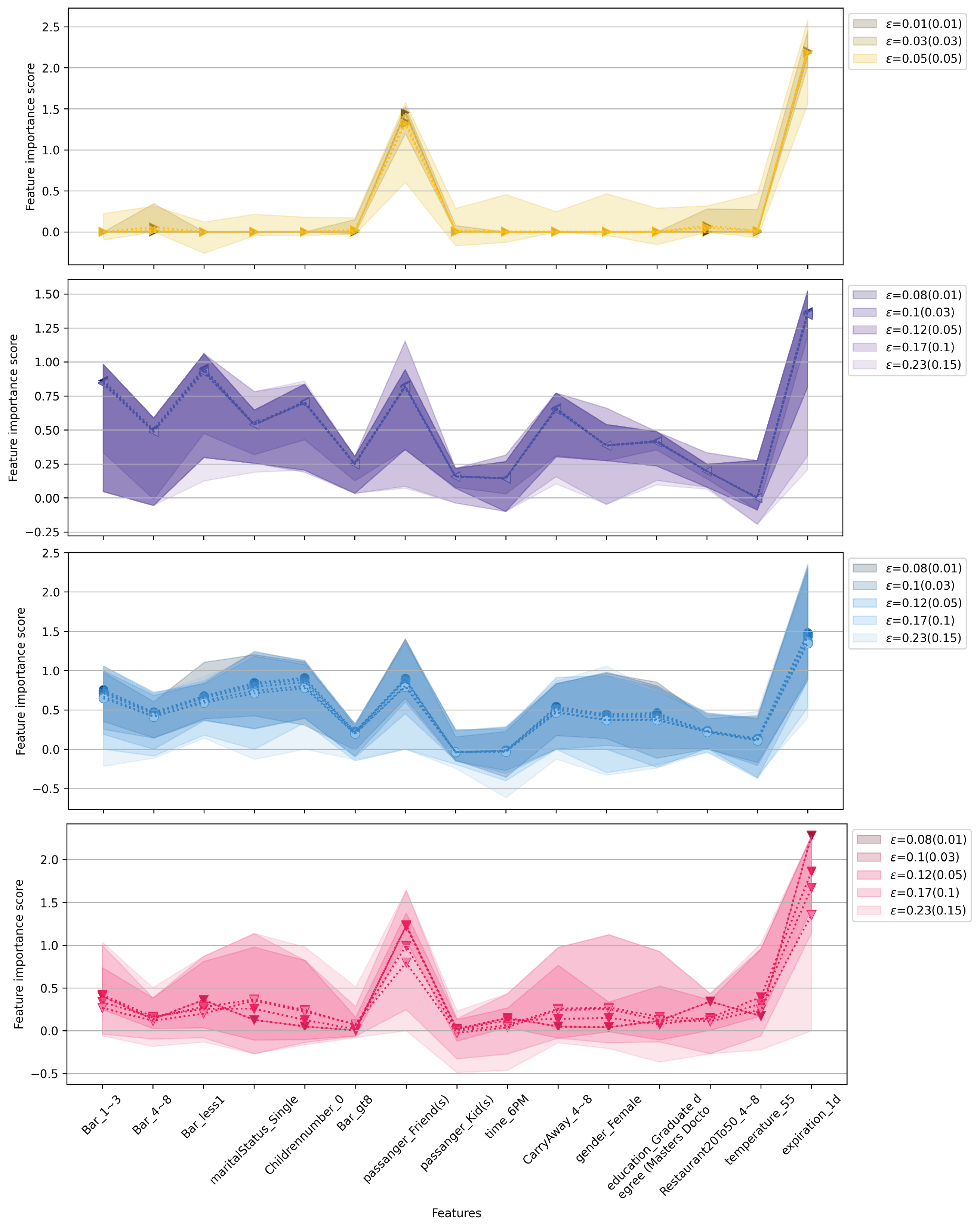}
     \end{subfigure}   
    \caption{Summary of detailed first-order feature attributions across different methods and epsilons on the dataset BAR, where the $x$-axis represents features and the $y$-axis displays feature importance scores. The legend includes epsilons relative to the reference model (in brackets) and to the optimal model identified in practice.} 
\label{fig:BAR-MR-ALL-E}
\end{figure}
\begin{figure}[h!]
     \centering
     \begin{subfigure}{\textwidth}
         \centering
         \includegraphics[width=\textwidth]{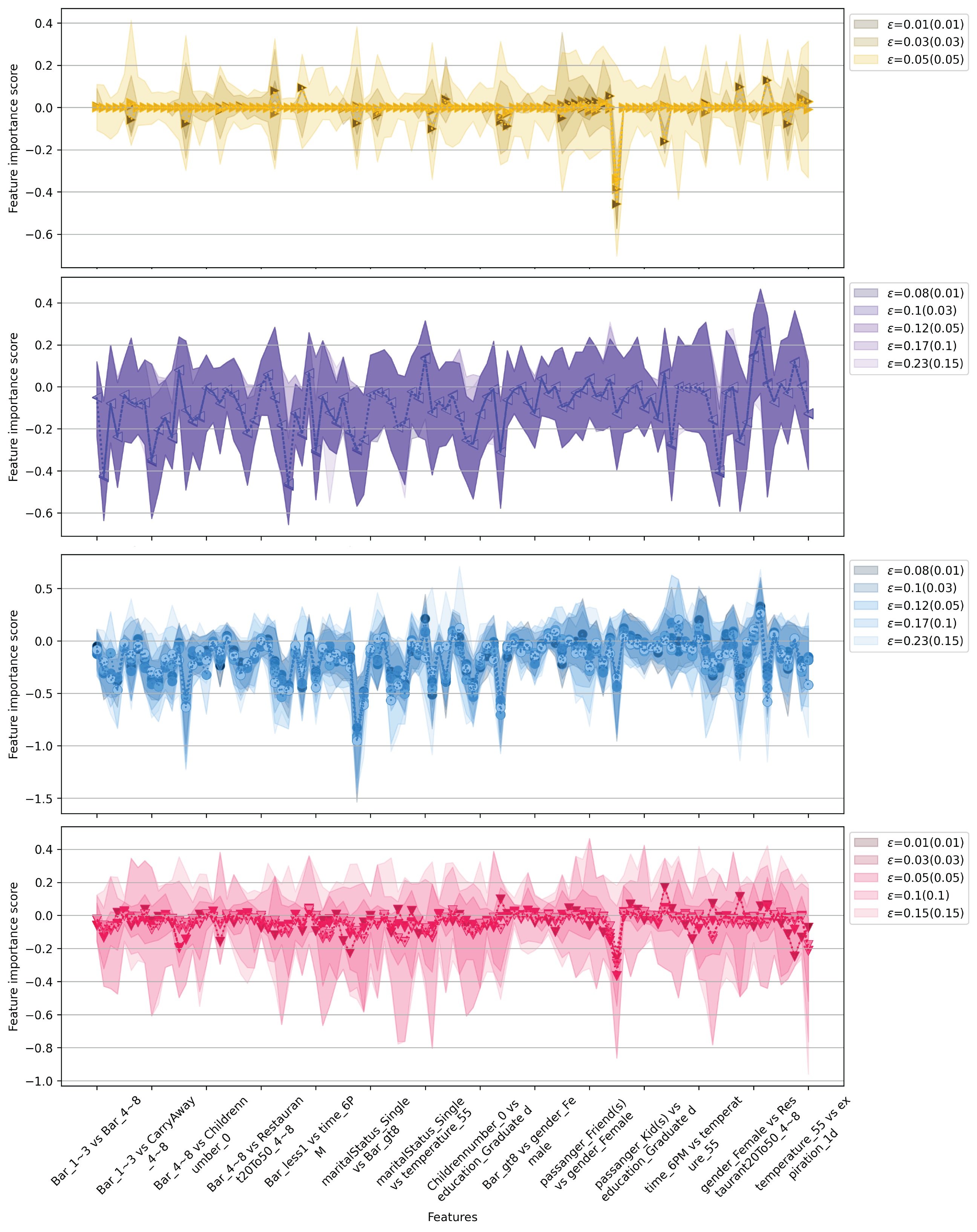}
     \end{subfigure}   
    \caption{SSummary of detailed second-order feature attributions across different methods and epsilons on the dataset BAR, where the $x$-axis represents features and the $y$-axis displays feature importance scores. The legend includes epsilons relative to the reference model (in brackets) and to the optimal model identified in practice.} 
\label{fig:BAR-FIS-ALL-E}
\end{figure}

\begin{figure}[h!]
     \centering
     \begin{subfigure}{\textwidth}
         \centering
         \includegraphics[width=.8\textwidth]{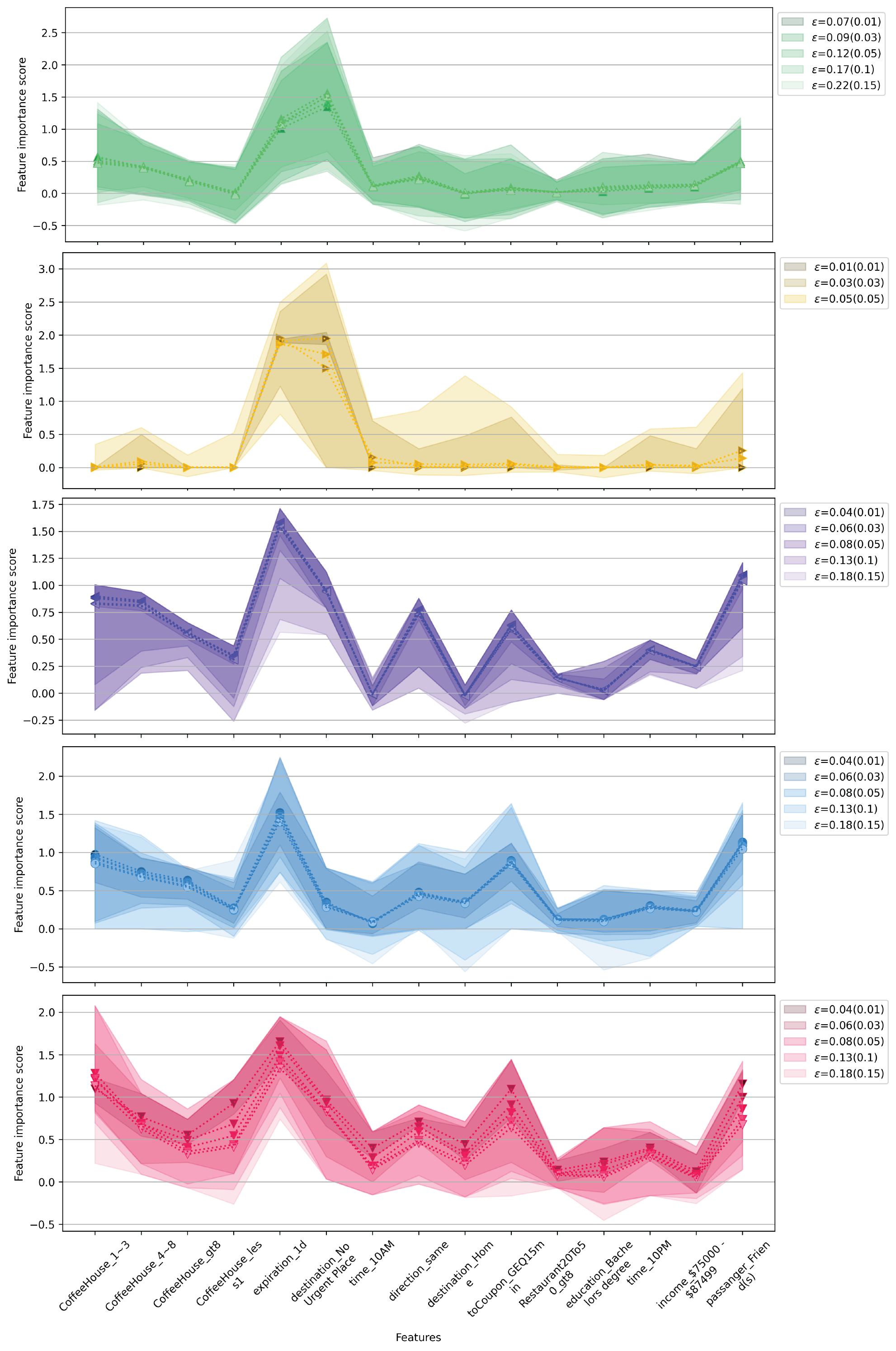}
     \end{subfigure}   
    \caption{Summary of detailed first-order feature attributions across different methods and epsilons on the dataset coffee house, where the $x$-axis represents features and the $y$-axis displays feature importance scores. The legend includes epsilons relative to the reference model (in brackets) and to the optimal model identified in practice.} 
\label{fig:COFFEE-MR-ALL-E}
\end{figure}
\begin{figure}[h!]
     \centering
     \begin{subfigure}{\textwidth}
         \centering
         \includegraphics[height=19cm]{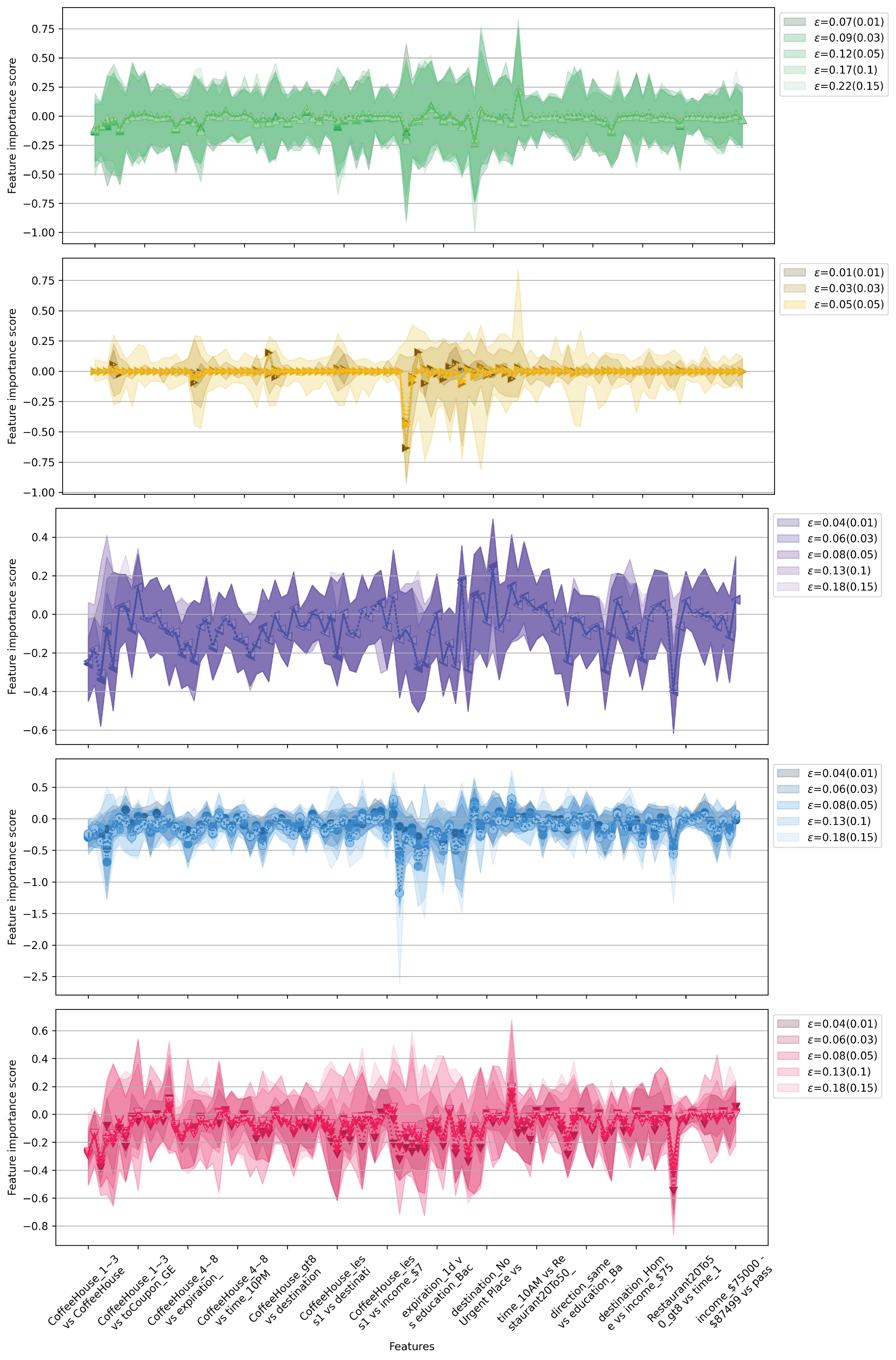}
     \end{subfigure}   
    \caption{Summary of detailed second-order feature attributions across different methods and epsilons on the dataset coffee house, where the $x$-axis represents features and the $y$-axis displays feature importance scores. The legend includes epsilons relative to the reference model (in brackets) and to the optimal model identified in practice.} 
\label{fig:COFFEE-FIS-ALL-E}
\end{figure}

\begin{figure}[h!]
     \centering
     \begin{subfigure}{\textwidth}
         \centering
         \includegraphics[width=.8\textwidth]{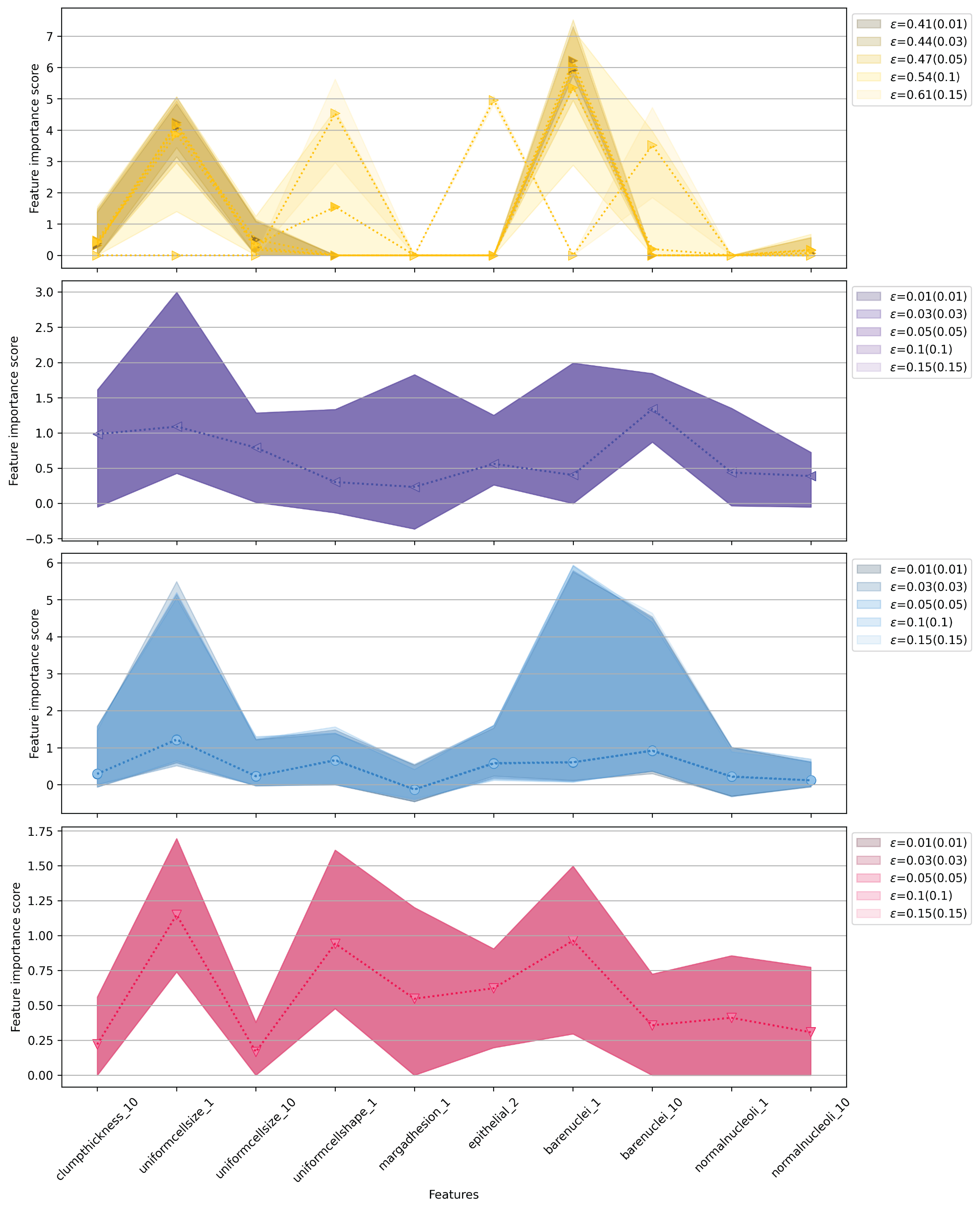}
         \label{fig:fig3}
     \end{subfigure}   
    \caption{Summary of detailed first-order feature attributions across different methods and epsilons on the dataset breast cancer, where the $x$-axis represents features and the $y$-axis displays feature importance scores. The legend includes epsilons relative to the reference model (in brackets) and to the optimal model identified in practice.} 
\label{fig:BREAST-MR-ALL-E}
\end{figure}
\begin{figure}[h!]
     \centering
     \begin{subfigure}{\textwidth}
         \centering
         \includegraphics[height=19cm]{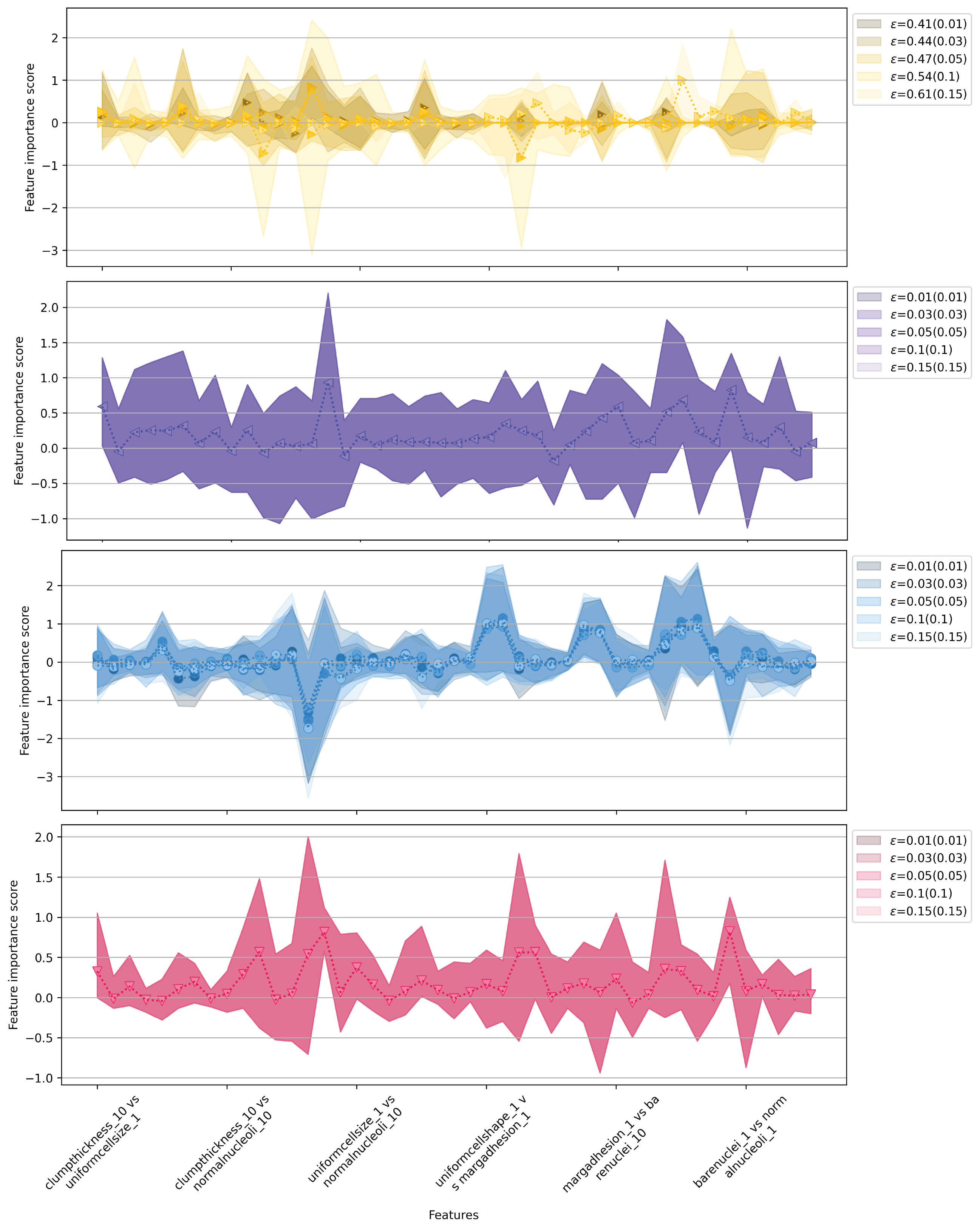}
         \label{fig:fig3}
     \end{subfigure}   
    \caption{Summary of detailed second-order feature attributions across different methods and epsilons on the dataset breast cancer, where the $x$-axis represents features and the $y$-axis displays feature importance scores. The legend includes epsilons relative to the reference model (in brackets) and to the optimal model identified in practice.} 
\label{fig:BREAST-FIS-ALL-E}
\end{figure}

\clearpage
\vskip 0.2in
\bibliography{sample}

\end{document}